\documentclass{ieeeaccess}
\usepackage{cite}
\usepackage{amsmath,amssymb,amsfonts}
\usepackage{algorithmic}
\usepackage{graphicx}
\usepackage{textcomp}
\usepackage{graphics} 
\usepackage{epsfig} 
\usepackage{multirow}
\usepackage{colortbl}
\usepackage{amssymb}
\usepackage{amsmath}
\usepackage{color}
\usepackage{booktabs}
\usepackage{bm}
\usepackage{colortbl}
\usepackage{array}
\usepackage{autobreak}
\usepackage{mathtools}
\usepackage{threeparttable}

\def\BibTeX{{\rm B\kern-.05em{\sc i\kern-.025em b}\kern-.08em
    T\kern-.1667em\lower.7ex\hbox{E}\kern-.125emX}}
\begin{document}
\history{Received January 14, 2021, accepted January 22, 2021, date of publication January 26, 2021, date of current version February 4, 2021.}
\doi{10.1109/ACCESS.2021.3054960}

\title{Motion Generation Using Bilateral Control-based Imitation Learning \\ with Autoregressive Learning}
\author{\uppercase{Ayumu Sasagawa}\authorrefmark{1}, \IEEEmembership{Student Member, IEEE},
\uppercase{Sho Sakaino}\authorrefmark{2,3}, \IEEEmembership{Member, IEEE}, \uppercase{and Toshiaki Tsuji}\authorrefmark{1}, \IEEEmembership{Senior Member, IEEE}}

\address[1]{Graduate School of Science and Engineering, Saitama University, Saitama 338-8570, Japan}
\address[2]{Department of Intelligent Interactive Systems, University of Tsukuba, Ibaraki 305-8577, Japan}
\address[3]{JST PRESTO, Saitama 332-0012, Japan}

\tfootnote{This work was supported in part by the JST PRESTO, Japan, under Grant JPMJPR1755.}
\markboth
{Sasagawa \headeretal: Motion Generation Using Bilateral Control-based Imitation Learning with Autoregressive Learning}
{Author \headeretal: Motion Generation Using Bilateral Control-based Imitation Learning with Autoregressive Learning}

\corresp{Corresponding author: Ayumu Sasagawa (e-mail: a.sasagawa.997@ms.saitama-u.ac.jp).}

\begin{abstract}
  Imitation learning has been studied as an efficient and high-performance method to generate robot motion.
  Specifically, bilateral control-based imitation learning has been proposed as a method of realizing fast motion.
  However, the learning approach of this method leads to the accumulation of prediction errors during the prediction process and may not generate desirable long-term behavior.
  Therefore, in this paper, we propose a method of autoregressive learning for bilateral control-based imitation learning to reduce the accumulation of prediction errors.
  A new neural network model for implementing autoregressive learning is also proposed.
  Three types of experiments are conducted to verify the effectiveness of the proposed method, where the method is shown to have improved performance over those of conventional approaches.
  Due to the structure and method of autoregressive learning employed by the developed model, the proposed method can generate desirable long-term motion for successful tasks and has a high generalization ability for environmental changes based on the human demonstrations of tasks.

\end{abstract}


\titlepgskip=-15pt

\maketitle

\section{INTRODUCTION}
\label{sec:introduction}
\PARstart{R}{obots} that can execute various tasks automatically are becoming an increasingly important focus of research in the field of robotics.
Recently, machine learning has exhibited high performance in various fields, including image recognition \cite{object detection}, machine translation \cite{attention}, expression recognition \cite{reviewer2-1}, human activity recognition \cite{human activity recognition}, and robotics \cite{reviewer2-2}--\cite{human-like}.
Also, approaches based on end-to-end learning for motion generation have also recently shown high performance \cite{end-to-end}--\cite{poke}.
Where approaches based on reinforcement learning require a considerable number of trials \cite{google},
end-to-end learning reduces the effort required for programming, and complex robotic motion can be easily generated.
Moreover, end-to-end learning methods have a high generalization ability for situation-related changes.
\par
Among these approaches, imitation learning (IL) and learning from demonstration have attracted attention as methods for efficiently learning robot motion \cite{imitation learning survey2}--\cite{relay policy}.
These learning-based methods use datasets derived from human demonstrations of tasks.
Yang {\it et al.} realized autonomous robot operations using neural network (NN) models \cite{ogata}, whereas another study proposed
a method that combines reinforcement learning and IL \cite{relay policy}.
IL using force information has also been proposed \cite{iit1}--\cite{force writing}, where force control improves the robustness against position fluctuations.
More specifically, force control increases the possibility of adapting to complex tasks that require force information and thus enables the accomplishment of a greater number of various tasks.
\cite{iit1}\cite{iit2} used haptic devices to collect force information during human demonstrations of tasks.
Rozo {\it et al.} realized cooperative work between a human and robot using a Gaussian mixture model \cite{cooperationIIT}, and Ochi {\it et al.} used NN models to integrate visual, position, and force information to perform tasks \cite{harada}.
In addition, \cite{force writing} used dynamic movement primitives to model human demonstrations of tasks and realized the automated task of writing letters.
A common problem with these approaches is that robot motion is extremely slow compared to that of humans.
\par
We previously proposed a bilateral control-based IL as a method that uses force information \cite{adachi}\cite{sasagawa}.
Bilateral control is a remote-control system that uses two robots, namely, a master and slave \cite{bilate}--\cite{bilate reviewer2}.
We employed bilateral control in a series of human demonstrations of tasks, where
a human operated the master robot, with the slave being teleoperated and conducting tasks within the workspace.
\Figure[t!](topskip=0pt, botskip=0pt, midskip=0pt)[width=6cm]{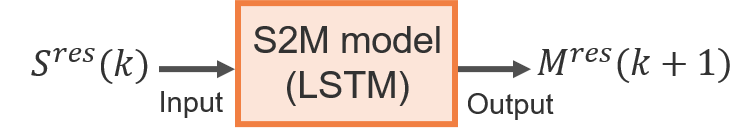}
{Network model of our bilateral control-based IL. \label{fig:S2M}}
As Fig.~\ref{fig:S2M} shows, the NN model for motion generation predicted the master state from the slave state.
The NN model included long short-term memory (LSTM) \cite{LSTM} to predict sequence data.
Here, $S$ and $M$ represent the slave and master, respectively, the superscript $res$ indicates the response values, and $k$ represents the step of the sequence data.
Our bilateral control-based IL can execute tasks requiring a force adjustment and realize fast motion that a conventional IL \cite{iit1}--\cite{force writing} cannot.
\Figure[t](topskip=0pt, botskip=0pt, midskip=0pt)[width=16cm]{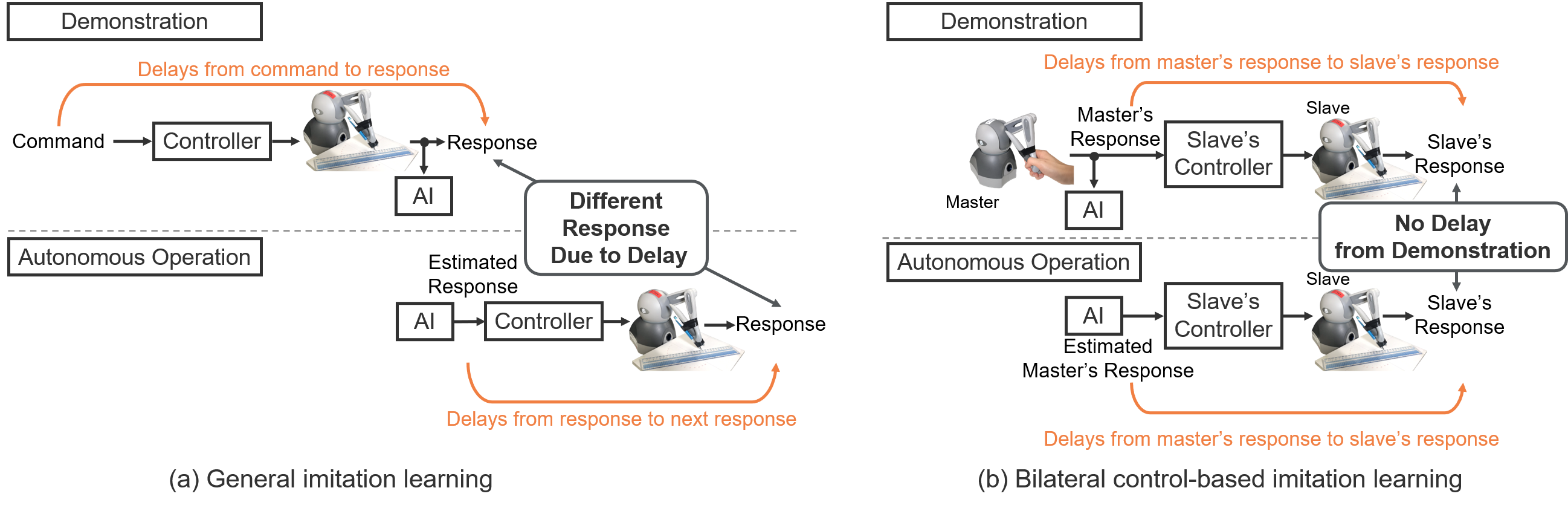}
{Overview of general IL and our bilateral control-based IL. In general, the delays caused during the demonstration and autonomous operation are different. Therefore, a general IL can realize only slow motion, thus ignoring delays.
In the bilateral control-based IL, the delays caused during the demonstration and autonomous operation are the same.
Thus, in our bilateral control-based IL, fast motion with delays can be achieved.\label{fig:delay}}
\Figure[t](topskip=0pt, botskip=0pt, midskip=0pt)[width=8.5cm]{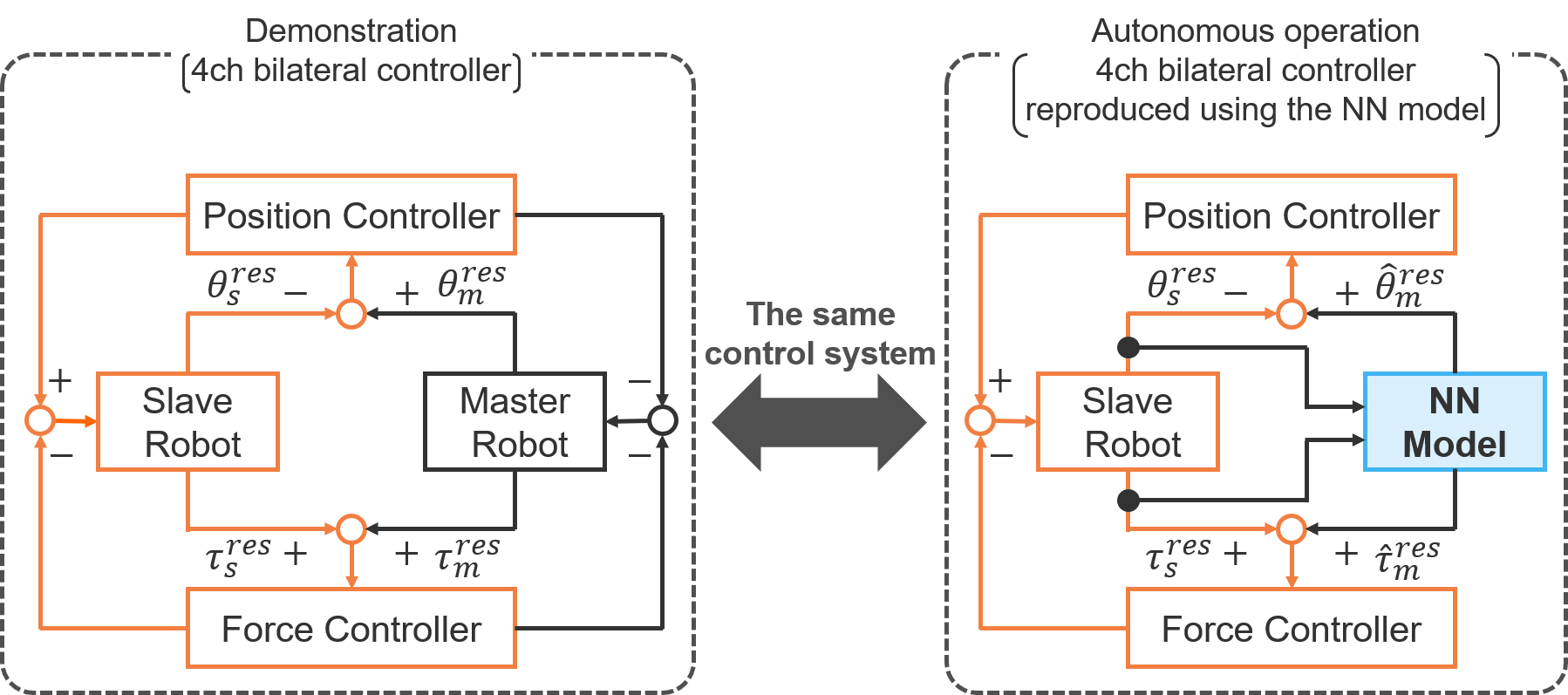}
{4ch bilateral controllers during the demonstrations and the autonomous operation.
The figure on the left side shows a 4ch bilateral controller.
This controller was used during the human demonstrations of tasks.
As the right side of the figure shows, the master robot and master's controllers were replaced with the NN model to virtually reproduce the 4ch bilateral controller during an autonomous operation.
With the method proposed in this study, both the master and slave response values were input to the NN model.
The same control system was applied during the human demonstrations of tasks and autonomous operation.
Note that the system including the slave robot and slave's controllers (orange‐colored lines) must be the same because the slave was used during the autonomous operation. \label{fig:4ch}}
As Fig.~\ref{fig:delay}(a) shows, in conventional IL, the response values collected during the demonstrations are given as the command values for autonomous operations, and delays among the command and the response values are not strictly considered.
However, in robot control, eliminating control delays is virtually impossible.
In addition, when performing tasks, including contact with the environment, delays due to physical interactions occur.
For this reason, a conventional IL can achieve only slow motion where delays can be ignored.
Based on the foregoing, the following two conditions must be satisfied to realize fast motion in the IL:
\begin{enumerate}
  \renewcommand{\labelenumi}{(\roman{enumi})}
  \item command values must be predicted during autonomous operation, {\it i.e.,} collected during the demonstrations,
  \item the same control system must be implemented during the demonstrations and autonomous operations.
\end{enumerate}
Our bilateral control-based IL can satisfy these two conditions for the following reasons.
First, in bilateral control, the command values of the slave are the response values of the master, and both of them can be measured independently.
As a result, the slave’s command values of the slave can be predicted during autonomous operations.
As Fig.~\ref{fig:delay}(b) shows, in our bilateral control-based IL, delays that occur during the demonstrations also occur during autonomous operations.
Second, as Fig.~\ref{fig:4ch} shows, in a bilateral control-based IL, the system is designed to reproduce bilateral control during an autonomous operation.
Thus, the control system of the slave can be the same during the demonstrations and autonomous operations.
In demonstrations using bilateral control, humans collect data while considering delays, {\it i.e.,} humans demonstrate skills to compensate for delays.
If the control system is different during demonstrations and autonomous operations, this compensation skill will be lost.
\par
Although our bilateral control-based IL can achieve a fast and dynamic motion, it does have a drawback. Specifically, the learning method of this approach is unsuitable for long-term predictions because the NN model is trained without autoregressive learning. This learning method is called teacher forcing \cite{techer forcing}.
When the NN model is trained using teacher forcing, if prediction errors occur during the prediction process, the errors will accumulate, and the robot will not realize a desirable behavior.
Autoregressive learning is a method to solve this problem, in which the output at the previous step is input to the model in the subsequent step. This method is called free running \cite{professor forcing}.
Because autoregressive learning predicts a series of motions continuously, the model is learned to minimize the total errors of the long-term prediction.
To implement autoregressive learning, the input and output of the model must be the same variables.
In general, the implementation of autoregressive learning is simple \cite{ogata closed} because the input and output of the model are the same variables, {\it i.e.,} response values.
By contrast, in our bilateral control-based IL, the output of the model cannot be used as the next input because the input and output of the model are different variables, {\it i.e.,} the response values of the slave and of the master (Fig.~\ref{fig:S2M}).
Therefore, we propose a model in which the input and output of the proposed model have both master and slave response values to implement autoregressive learning in a bilateral control-based IL.
In summary, the main contributions of this paper are as follows:
\begin{enumerate}
  \renewcommand{\labelenumi}{(\roman{enumi})}
  \item A new NN model for autoregressive learning in our bilateral control-based IL is proposed.
  \item The proposed model improves the ability to generate long-term behavior, and it leads to improve the adaptability to environmental changes.
\end{enumerate}
\par
In this study, the proposed model was compared with conventional models.
For the experiments, three tasks were conducted to clarify the effectiveness of the proposed method.
The success rates of the tasks were used to evaluate the performance of the method.
In all experiments, the proposed method showed an excellent performance equal to or greater than that of previous conventional methods.
\par
The remainder of this paper is organized as follows.
Section \ref{sec:control system} introduces the control system and bilateral control, and Section \ref{sec:method} describes the method of the bilateral control-based IL. Section \ref{sec:nn model} describes autoregressive learning and the NN models for the proposed method and previous conventional methods.
Section \ref{sec:experiment} describes the experiments and presents the results of the three tasks. Section \ref{sec:conclusion} provides concluding remarks regarding this study and discusses areas of future research.

\section{CONTROL SYSTEM}
\label{sec:control system}
\subsection{Robot}
\label{subsec:manipulater}
\Figure[t](topskip=0pt, botskip=0pt, midskip=0pt)[width=4cm]{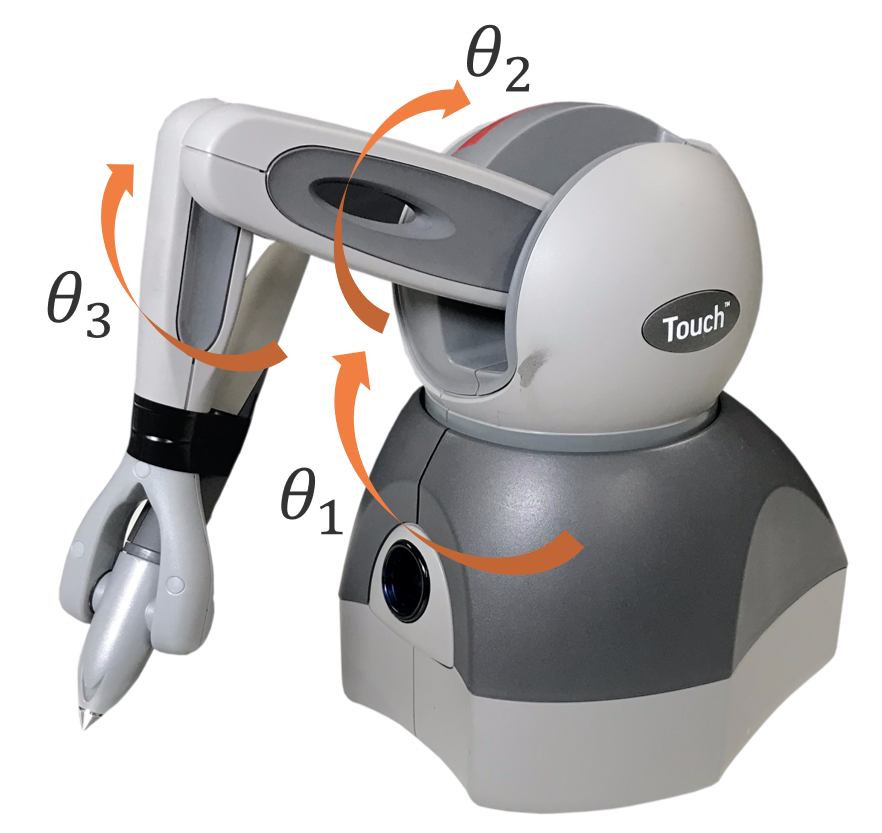}
{Robot (Touch${}^\text{\tiny TM}$).\label{fig:phantom}}
Two Touch${}^\text{\tiny TM}$, which are haptic devices manufactured by 3D Systems, were used in the experiments.
Two robots were used as master and slave, respectively.
The robots had three-degrees-of-freedom (DOF), as shown in Fig.~\ref{fig:phantom}.
The robots could measure only the joint angles $\theta_1$, $\theta_2$, and $\theta_3$ with the encoders.
Here, the subscripted numbers represent each joint shown in Fig.~\ref{fig:phantom}.
\subsection{Bilateral control}
\label{subsec:bilateral control}
Bilateral control is a type of remote-control system that uses two robots, {\it i.e.,} a master and slave \cite{bilate}--\cite{bilate reviewer2}.
In this study, 4ch bilateral control \cite{4ch bilate 1}\cite{4ch bilate 2} was used from among various types of bilateral control because it offers the highest performance and excellent operability, and the slave and master consist of both position and force controllers.
Therefore, 4ch bilateral control is suitable for IL \cite{sasagawa}.
In bilateral control, when the operator operates the master, the slave is teleoperated.
The control goal is to synchronize the position and satisfy the law of action and reaction forces between the two robots.
The reaction force caused by the contact between the slave and environment is presented to the master.
Thus, the operator can feel the interactions between the slave and environment.
The control law of the 4ch bilateral control is expressed through the following equations using the angle response values $\theta^{res}$ and torque response values $\tau^{res}$ of the robots.
A block is given on the left side of Fig.~\ref{fig:4ch}.
Note that the subscripts $s$ and $m$ represent the slave and master, respectively, and the superscript $res$ represents the response values.
\begin{eqnarray}
  \theta^{res}_{m}-\theta^{res}_{s} & = & 0, \label{eq1} \\
  \tau^{res}_{m}+\tau^{res}_{s} & = & 0. \label{eq2}
\end{eqnarray}
\subsection{Controller}
\label{subsec:controller}
\Figure[t](topskip=0pt, botskip=0pt, midskip=0pt)[width=6.5cm]{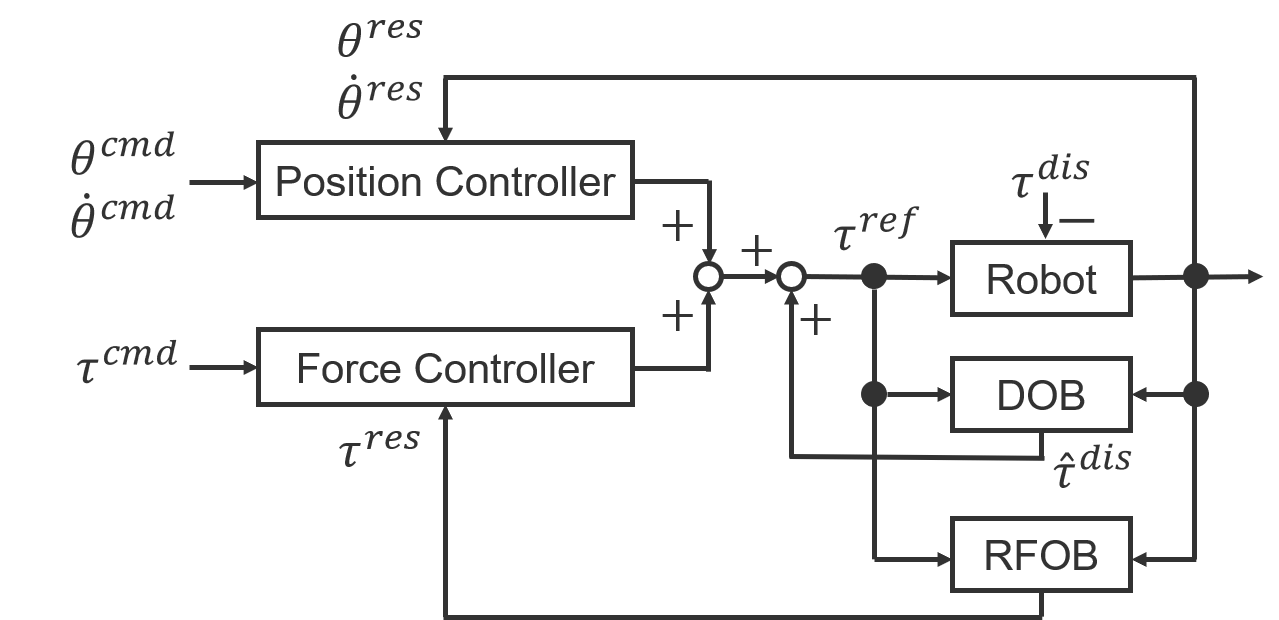}
{Controller.\label{fig:controller}}
The control system consisted of position and force controllers, as shown in Fig.~\ref{fig:controller}.
Here, $\theta$, $\dot{\theta}$, and $\tau$ represent the joint angles, angular velocity, and torque of each joint, respectively, the superscripts $res$, $cmd$, and $ref$ indicate the response, command, and reference values, respectively. In addition, $\theta^{res}$ was measured by the encoders of the robots, and $\dot{\theta}^{res}$ was calculated using pseudo-differentiation.
The disturbance torque $\tau^{dis}$ was estimated using a disturbance observer (DOB) \cite{dob} as $\hat{\tau}^{dis}$.
Furthermore, a reaction force observer (RFOB) \cite{rfob} was used to calculate the reaction force $\tau^{res}$.
Details of the RFOB are provided in Section~\ref{subsec:identification}.
The position controller also included proportional and derivative controllers, where the force controller comprised a proportional controller.
The torque reference values $\tau^{ref}$ of the slave and master were calculated as follows:
\begin{eqnarray}
    \tau^{ref}_{m} & = & - \frac{J}{2}(K_{p} + K_{d}s) (\theta^{res}_{m} - \theta^{res}_{s}) \nonumber \\ &&- \frac{1}{2}K_{f} (\tau^{res}_{m}+\tau^{res}_{s}), \label{eq3} \\
    \tau^{ref}_{s} & = &   \frac{J}{2}(K_{p} + K_{d}s) (\theta^{res}_{m} - \theta^{res}_{s}) \nonumber \\ &&- \frac{1}{2}K_{f} (\tau^{res}_{m}+\tau^{res}_{s}), \label{eq4}
\end{eqnarray}
where $s$ represents the Laplace operator, $J$ is the inertia, and $K_p$, $K_d$, and $K_f$ represent the position, velocity, and force control gain, respectively.
The gain values and cut-off frequencies used in the experiments are listed in Table~\ref{table:parameter}.
\par
Also, robot dynamics are represent as follows:
\begin{eqnarray}
  J_{1}\ddot{\theta}^{res}_{1} & = & \tau^{ref}_{1} - \tau^{dis}_{1} - D \dot{\theta}^{res}_{1},\label{eq:dob1} \\
  J_{2}\ddot{\theta}^{res}_{2} & = & \tau^{ref}_{2} - \tau^{dis}_{2} - G_{1} \cos{\theta^{res}_{2}} - G_{2} \sin{\theta^{res}_{3}},\label{eq:dob2} \\
  J_{3}\ddot{\theta}^{res}_{3} & = & \tau^{ref}_{3} - \tau^{dis}_{3} - G_{3} \sin{\theta^{res}_{3}}.\label{eq:dob3}
\end{eqnarray}
where $D$ and $G$ represent the friction compensation and gravity compensation coefficients, respectively, and the subscripts represent the different joints of the robots.
Also, $\ddot{\theta}$ means angular acceleration.
The off-diagonal term of the inertia matrix was ignored because it was negligibly small. 
\begin{table}[t]
  \tabcolsep = 3pt
  \begin{center}
  \caption{Gains and identified system parameters \protect\linebreak for the robot controller}
  \label{table:parameter}
  \begin{tabular}{lcc}
      \hline\hline
      & Parameter & Value  \\
      \hline
      \arrayrulecolor{black}
      $J_{1}$            & Joint 1's inertia [$\rm{mkgm^2}$]   & 2.55 \\
      $J_{2}$            & Joint 2's inertia [$\rm{mkgm^2}$]   & 4.30 \\
      $J_{3}$            & Joint 3's inertia [$\rm{mkgm^2}$]   & 1.12 \\
      $G_{1}$                       & Gravity compensation coefficient 1 [$\rm{mNm}$]   & 79.0 \\
      $G_{2}$                       & Gravity compensation coefficient 2 [$\rm{mNm}$]   & 55.0\\
      $G_{3}$                       & Gravity compensation coefficient 3 [$\rm{mNm}$]   & 33.0\\
      $D$                         & Friction compensation coefficient[$\rm{mkgm^2/s}$]   & 4.55\\
      \hline
      $K_p$                       & Position feedback gain              & 121 \\        
      $K_d$                       & Velocity feedback gain                & 22.0 \\
      $K_f$                       & Force feedback gain              & 1.00 \\
      $g$ &Cut-off frequency of pseudo differentiation [$\rm{rad/s}$]& 40.0 \\
      $g_{DOB}$ &Cut-off frequency of DOB [$\rm{rad/s}$]& 40.0 \\
      $g_{RFOB}$ &Cut-off frequency of RFOB [$\rm{rad/s}$]& 40.0 \\
      \hline\hline
  \end{tabular}
  \end{center}
\end{table}

\subsection{System identification}
\label{subsec:identification}
The parameters of the control system were identified based on \cite{yamazaki}. 
Friction $D$ and gravity $G$ were identified under free motion, assuming $\tau^{dis} = 0$. 
The DOB calculated the estimated disturbance torque $\hat{\tau}^{dis}$ as follows:
\begin{eqnarray}
  \hat{\tau}^{dis} & = & \tau^{ref} - J\ddot{\theta}^{res}.\label{eq:dob4}
\end{eqnarray}
When the RFOB is used, each force can be measured without using a force sensor, and the torque response values of each joint were calculated as follows:
\begin{eqnarray}
  \tau^{res}_{1} & = & \hat{\tau}^{dis}_{1} - D \dot{\theta}^{res}_{1}, \label{eq:dob5} \\
  \tau^{res}_{2} & = & \hat{\tau}^{dis}_{2} - G_{1} \cos{\theta^{res}_{2}} - G_{2} \sin{\theta^{res}_{3}}, \label{eq:dob6} \\
  \tau^{res}_{3} & = & \hat{\tau}^{dis}_{3} - G_{3} \sin{\theta^{res}_{3}}. \label{eq:dob7}
\end{eqnarray}
All identified parameters used in the experiment are listed in Table~\ref{table:parameter}.

\section{METHOD OF BILATERAL CONTROL-BASED IMITATION LEARNING}
\label{sec:method}
In our experiments, the robots learned behaviors from human demonstrations of tasks and then conducted the tasks autonomously.
In the demonstrations, the desired tasks were conducted using 4ch bilateral control.
A human operated the master, and the slave performed the tasks in the given workspace.
The joint angles, angular velocity, and torque response values of the two robots were used as the dataset for model training.
Both the control and data-saving cycles were 1~ms long.
\par
The NN model was then trained using the dataset derived from the demonstrations.
The NN model consisted of LSTM and fully connected layers to learn the time-series data.
The number of LSTM layers was set for each task by trial and error.
Details are described in Section~\ref{sec:experiment}.
The activation functions for the LSTM and fully connected layers were a hyperbolic tangent function and identity mapping, respectively.
Essentially, the model was trained to input the state at time $t$ and output the state at time $t+20$~ms.
Whether either the input or output were the master or slave state depended on each model described in Section~\ref{sec:nn model}.
The state consisted of the joint angles, angular velocity, and torque response values.
The loss function is the mean squared error (MSE) between the model output values and true values of the dataset.
The model was learned to minimize the loss function with Adam optimization \cite{adam}.
The dataset values were normalized to [0, 1] before the input to the model.
\par
Finally, the trained model generated the motion, and the robot autonomously conducted the tasks.
The control system was designed to reproduce 4ch bilateral control during the autonomous operation.
The joint angle, angular velocity, and torque response values of the slave were measured in real time and input to the learned model.
The command values predicted by the model were normalized before the input to the slave controller.
Note that the prediction and control cycles of the model and robot were 20 and 1~ms, respectively.

\section{NEURAL NETWORK MODEL}
\label{sec:nn model}
\subsection{Autoregressive learning}
\label{subsec:sutoregressive learning}
\Figure[t](topskip=0pt, botskip=0pt, midskip=0pt)[width=15cm]{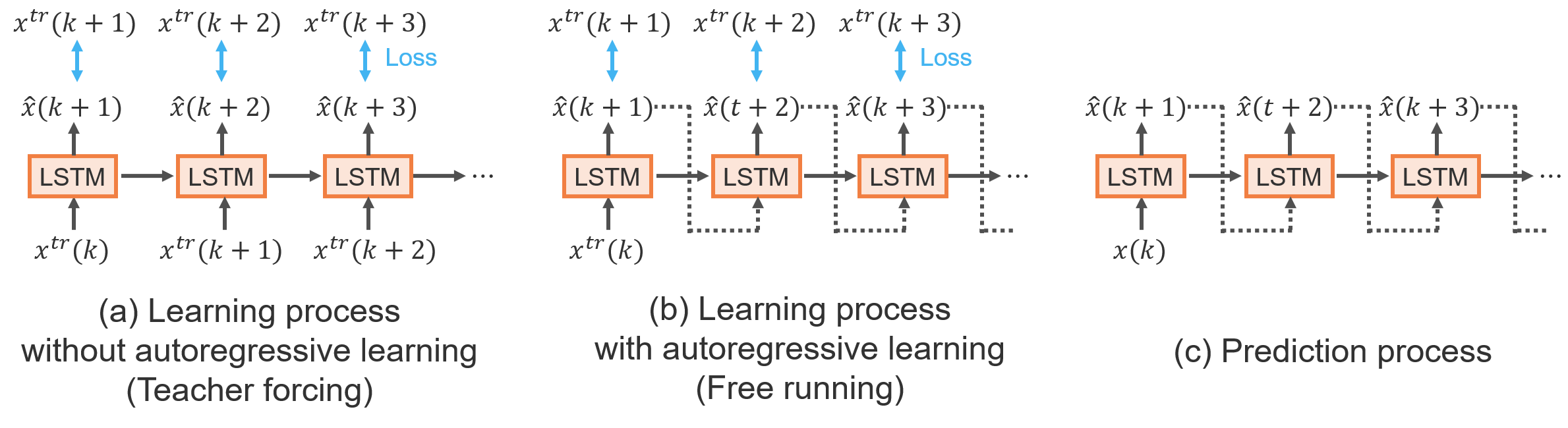}
{Learning and prediction methods using LSTM with and without autoregressive learning.\label{fig:autoregression}}
Fig.~\ref{fig:autoregression} shows the LSTM model developed in the time direction.
Here, $x$ represents an arbitrary value used for the input and output, the superscript $tr$ represents teacher data, and $\hat{\bigcirc}$ represents the predicted values of the model.
Fig.~\ref{fig:autoregression}(a) shows the learning method without autoregressive learning.
With this method, the teacher data are input at each step during the learning process, {\it i.e.,} the input values are completely unaffected by the prediction in the previous steps.
More precisely, the output error used to train the NN model is based only on a one-step prediction.
\par
However, as Fig.~\ref{fig:autoregression}(c) shows, the model's prediction values are used as the next input during the prediction process.
Therefore, without autoregressive learning, the input probability distributions are different between the learning and prediction processes.
If prediction errors occur because of this difference, they will accumulate step by step.
Therefore, although the model is learned with high accuracy during the learning process, it cannot generate desirable behavior during the prediction process.
This problem similarly occurs in the field of natural language processing when using a recurrent NN \cite{exposure bias1}--\cite{exposure bias4}.
It is known that the gap in input distributions can be filled by autoregressive learning, which uses the model's predicted values as input during the learning process, as shown in Fig.~\ref{fig:autoregression}(b).
In addition, during autoregressive learning, prediction errors accumulate step by step during the learning process, and the model is optimized to reduce these errors.
Therefore, the model is learned to behave properly throughout the continuous-time series.
Autoregressive learning prevents error accumulation during the prediction process, and the model is more likely to generate desirable behavior in the long-term to execute tasks.
\par
In a conventional bilateral control-based IL \cite{adachi}\cite{sasagawa}, autoregressive learning cannot be implemented.
This is because the input and output of the model are different variables, {\it i.e.,} the slave and master have different response values (Fig.~\ref{fig:S2M}).
In this study, the SM2SM model is proposed to solve this problem.
In addition, the performances of three models are compared with that of the proposed method.
Summary information about the different models is provided in Table~\ref{table:nn model}.
The general IL \cite{iit1}--\cite{force writing} predicts the next response values from the current ones.
Therefore, the S2S model that predicts the next slave state from the current one is used as a comparison method that replicates the general IL.
The S2M model is used as the conventional bilateral control-based IL \cite{adachi}\cite{sasagawa} and the SM2SM model is used as the proposed method.
Details of each model are provided in the following sections.
\begin{table*}[t]
  \begin{center}
  \caption{Details of the NN model}
  \label{table:nn model}
  \begin{tabular}{|l||c|c|c|}
      \hline
           & \multicolumn{2}{c|}{Neural network model} &\\
      \cline{2-3}
      Model  & Input & Output &Autoregressive learning \\
      \arrayrulecolor{black}
      \hline
      S2S-w/o-AR  & Slave (9~dims.) & Slave (9~dims.) & - \\
      S2S-AR & Slave (9~dims.) & Slave (9~dims.) & \checkmark \\
      S2M-w/o-AR & Slave (9~dims.) & Master (9~dims.) & - \\
      SM2SM-w/o-AR & Slave and master (18~dims.) & Slave and master (18~dims.) & - \\
      SM2SM-AR (Proposed model)       & Slave and master (18~dims.) & Slave and master (18~dims.) & \checkmark \\
      \hline
    \end{tabular}
  \end{center}
\end{table*}
\subsection{S2S model (Conventional model)}
\label{subsec:S2S}
\Figure[t](topskip=0pt, botskip=0pt, midskip=0pt)[width=6cm]{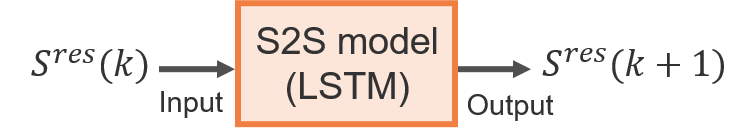}
{Network version of the S2S model (conventional method).\label{fig:S2S}}
As Fig.~\ref{fig:S2S} shows, the S2S model predicts the next slave state from the current one.
The input and output consist of the joint angles, angular velocity, and torque response values of the slave with three DOFs. In other words, the S2S model has nine inputs and nine outputs.

\subsubsection{Learning}
\label{subsubsec:training S2S}
During the learning process, the slave's response values were input, and the slave's response values 20~ms later were output.
The S2S model was trained without or with autoregressive learning.
The cases without and with autoregressive learning were called S2S-w/o-AR and S2S-AR, respectively.
The number of autoregressive steps was set to 10 to allow the prediction errors to converge quickly.
In other words, every 10 steps, the values of the training dataset were input instead of the predicted values from the previous step.

\subsubsection{Autonomous operation}
\label{subsubsec:operation S2S}
The model predicted the response values of the slave.
The predicted values of the model were used as the command values of the slave.

\subsection{S2M model (Conventional model)}
\label{subsec:S2M}
\Figure[t](topskip=0pt, botskip=0pt, midskip=0pt)[width=18cm]{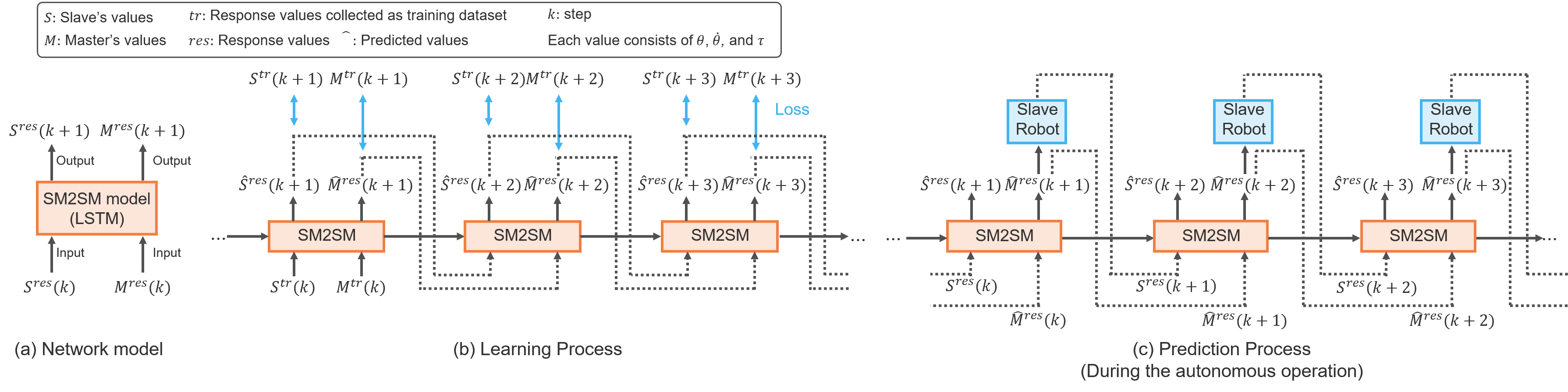}
{Learning process and prediction process of the SM2SM model (proposed method).\label{fig:SM2SM}}
As Fig.~\ref{fig:S2M} shows, the S2M model predicted the next state of the master from the current state of the slave.
The input consisted of the joint angles, angular velocity, and torque response values of the slave with three DOFs.
The output consisted of these response values of the master with three DOFs.
Therefore, the S2M model had nine inputs and nine outputs.

\subsubsection{Learning}
\label{subsubsec:training S2M}
During the training, the response values of the slave were input, and the model output the response values of the master 20~ms after the input was applied.
In the case of the S2M model, the model was trained without autoregressive learning because it could not be implemented.
The S2M model without autoregressive learning is called S2M-w/o-AR.

\subsubsection{Autonomous operation}
\label{subsubsec:operation S2M}
The model predicted the response values of the master; these predicted values were used as the command values of the slave.

\subsection{SM2SM model (Proposed model)}
\label{subsec:SM2SM}
SM2SM was the proposed model applied to adapt autoregressive learning to a bilateral control-based IL.
As Fig.~\ref{fig:SM2SM}(a) indicates, the SM2SM model predicted the next state of the slave and master from their respective current states.
In contrast to the S2M model, the input and output of the SM2SM model consisted of both slave and master states.
Therefore, autoregressive learning could be implemented.
In addition, the structure of this model was shown to have another advantage.
In bilateral control-based IL, the relationship between the slave and master must be learned accurately to reproduce the demonstrations.
During the learning of the S2M model, only one relationship was used, which was the prediction from the slave to the master state.
By contrast, in the SM2SM model, four relationships were used: predictions from the slave to the slave state, from the slave to the master state, from the master to the slave state, and from the master to the master state. These improved the learning of the dynamics of bilateral control.
Based on the foregoing, because interactions between master and slave robots could be implicitly learned by the SM2SM model, it was expected that the SM2SM model would be a suitable model for bilateral control-based IL.
The input and output for the SM2SM model consisted of the joint angles, angular velocity, and torque response values of the slave and master with three DOFs. In other words, the SM2SM model had 18 inputs and 18 outputs.

\subsubsection{Learning}
\label{subsubsec:training SM2SM}
An overview of the learning process of the SM2SM model is shown in Fig.~\ref{fig:SM2SM}(b).
During the learning process, the response values of the slave and master were input, and the response values of the slave and master 20~ms later were output.
The SM2SM model was learned without or with autoregressive learning.
The cases without and with autoregressive learning were called SM2SM-w/o-AR and SM2SM-AR, respectively.
In this study, the number of autoregressive steps was set to 10 to ensure that the prediction errors converged quickly.

\subsubsection{Autonomous operation}
\label{subsubsec:operation SM2SM}
Overview during the autonomous operation is shown in Fig.~\ref{fig:SM2SM}(c).
The slave state among the input to the model was the slave response values measured in real time.
By contrast, the master state among the inputs of the model was that predicted by the model one step before.
The states of the master predicted by the model were used as the command values of the slave.

\section{EXPERIMENT}
\label{sec:experiment}
In the experiment, three types of tasks were conducted to clarify the effectiveness of the proposed method.
The following abilities were verified in the three tasks:
\begin{enumerate}
  \renewcommand{\labelenumi}{(\roman{enumi})}
  \item Adaptability to environmental changes (confirmed in experiment 1 [\ref{subsec:experiment1}]).
  \item Ability to realize fast motion (confirmed in experiment 2 [\ref{subsec:experiment2}]).
  \item Ability to generate desirable long-term behavior (confirmed in experiment 3 [\ref{subsec:experiment3}]).
\end{enumerate}
Three types of NN models were compared for each experiment.
Only the S2M model was conducted without autoregressive learning, whereas the S2S and SM2SM models were compared with and without autoregressive learning, thus resulting in five types of models being compared.
The success rates of the tasks verified the performance of each model.

\subsection{Experiment 1 (Drawing a line using a pen and ruler)}
\label{subsec:experiment1}
\subsubsection{Task design}
\label{subsubsec:task1}
\Figure[t](topskip=0pt, botskip=0pt, midskip=0pt)[width=15cm]{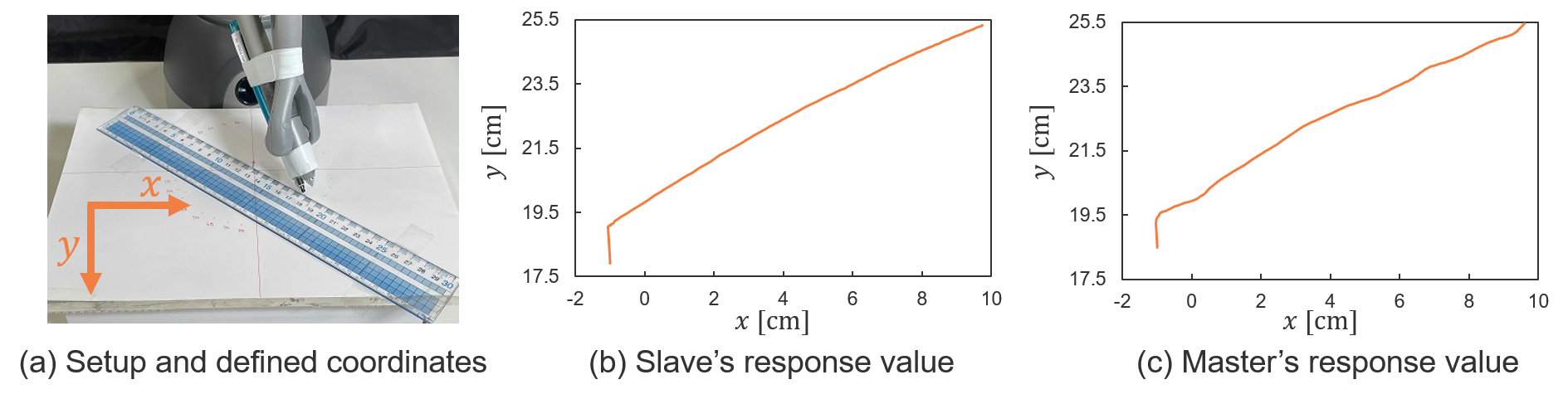}
{Setup and training data of the drawing task when drawing a 40 degree line:
(b) The response values of the slave did not include large fluctuations.
(c) By comparison, the response values of the master included large fluctuations (Fig.~c).
\label{fig:ruler_consideration}}
Fig.~\ref{fig:ruler_consideration}(a) shows the setup of this experiment.
A mechanical pencil was fixed to the slave.
Initially, the slave moved from the initial position toward the ruler.
After touching the ruler, the slave then drew a straight line to the right along the ruler.
The goal of this task was to draw lines according to various inclinations.
As Fig.~\ref{fig:ruler_illust}(a) shows, the inclination was defined by the angle at which the ruler was rotated around the point where the pen first contacted the ruler.
Zero degrees is represented by the "reference line" in the figure.
To succeed in this task, a proper adjustment of the contact force between the pen and ruler or paper was required.
In addition, adaptability to unknown inclinations or unknown positions of the ruler was required.

\subsubsection{Human demonstrations and dataset for learning}
\label{subsubsec:dataset1}
\Figure[t](topskip=0pt, botskip=0pt, midskip=0pt)[width=10cm]{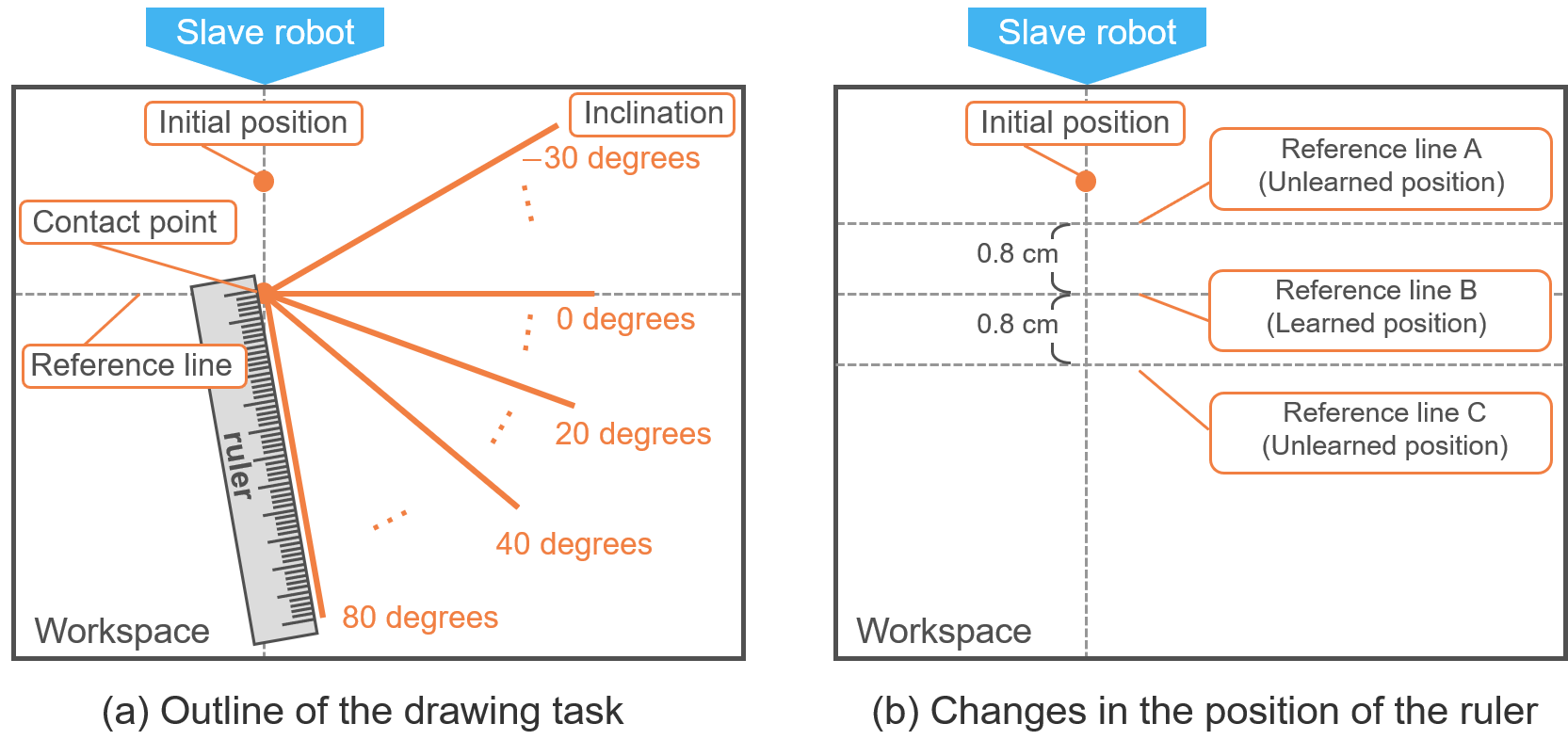}
{Outline of the drawing task.\label{fig:ruler_illust}}
We collected data using ruler inclinations of zero, 20, and 40 degrees, as shown Fig.~\ref{fig:ruler_illust}-(a).
Eight trials were conducted for each inclination, and the total number of trials was 24.
The time for one trial was 3~s.
The slave began moving from the initial position and, within 3~s, drew a line of 5~cm or longer along the edge of the ruler.

\subsubsection{Nural network architecture}
\label{subsubsec:nn1}
The NN model consisted of six LSTM layers, followed by a fully connected layer.
The unit size of all layers was 50.
The mini-batch consisted of 100 random sets of 150 time-sequential samples corresponding to 3~s.
The loss graph for each model is given as Fig.~\ref{fig:lossgraph}.
As Figs.~\ref{fig:lossgraph}(b) and (e) show, models with autoregressive learning required many learning epochs to converge the loss function.
Note that the performance of the autonomous operation did not necessarily depend on the value of the loss function.
Although the loss value was useful for roughly identifying the high-performance model, we identified the optimal number of learning epochs by comparing the performance of the autonomous operation in trial and error.
\Figure[t](topskip=0pt, botskip=0pt, midskip=0pt)[width=8cm]{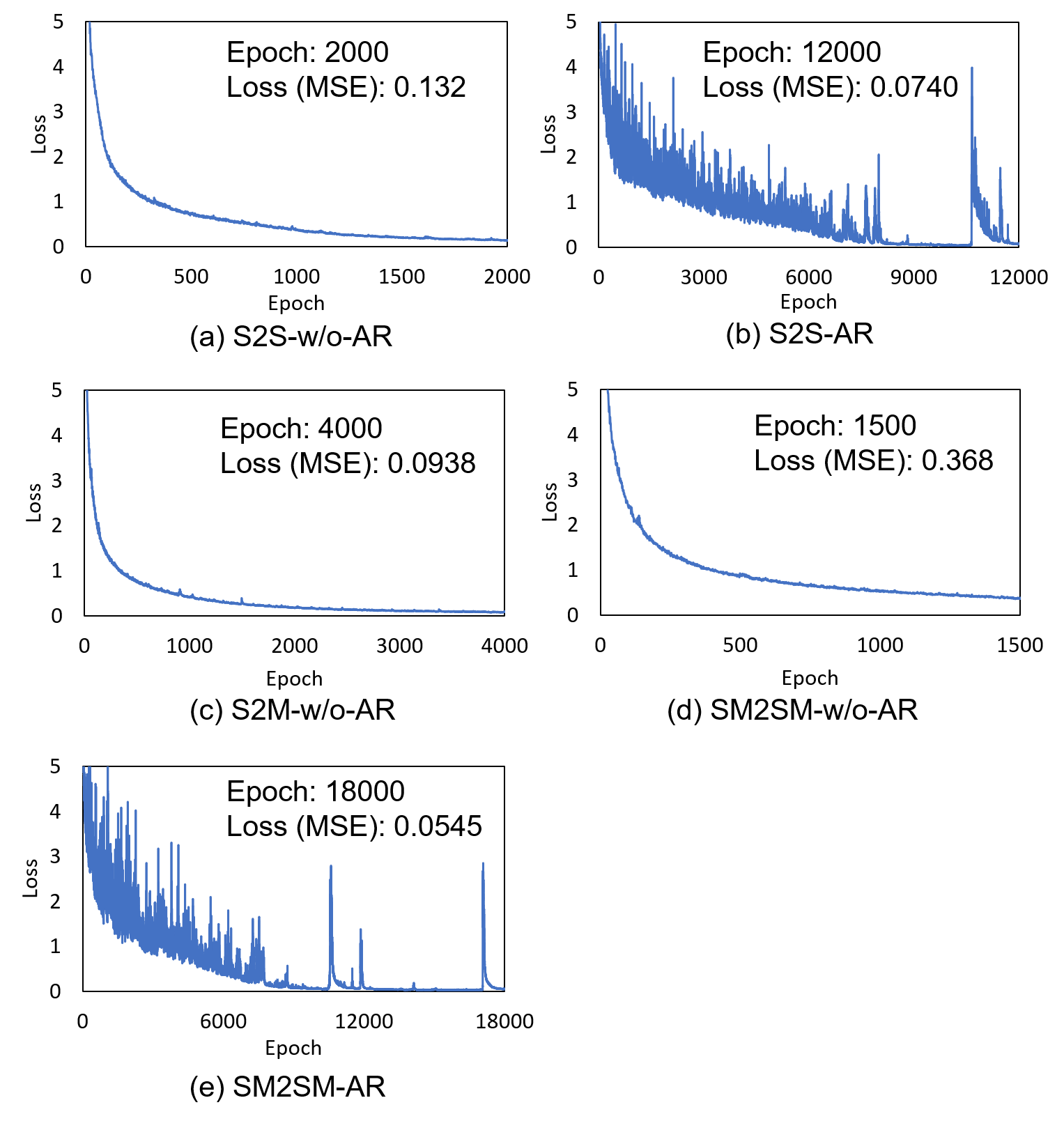}
{Relationship between loss and epoch number in the drawing task. The vertical axis shows the prediction loss calculated by MSE. The maximum value was set to 5 to enlarge the graph. The horizontal axis shows the number of learning epochs, and the maximum value of each model is the number of epochs used for the validation. In addition, the epoch and loss in the graph represent the number of epochs used for the validation and the value of the MSE at each epoch, respectively.\label{fig:lossgraph}}

\subsubsection{Task validation}
\label{subsubsec:validation1}
To verify the autonomous operation, the performance of the ruler inclinations from $-$30 to 80 degrees was verified every 10 degrees.
Success was defined by those cases in which the robot drew a line of 5 cm or longer along the edge of the ruler.
Verification was conducted through three trials for the inclination of each ruler.
In addition, the performance was validated when the ruler's position was shifted back and forth.
Here, the position of the ruler was defined based on the point where the pen first contacted the ruler.
As Fig.~\ref{fig:ruler_illust}(b) shows, the validation was conducted by shifting 0.8~cm back and forth from the learned position.
The learned position was ``reference line B,'' and the unlearned positions were ``reference line A'' and ``reference line C.''

\subsubsection{Experimental results}
\label{subsubsec:result1}

\begin{table*}[t]
  \begin{center}
    \begin{threeparttable}[t]
    \caption{Success rates of the drawing task}
    \label{table:result_ruler}
    \scriptsize
    \begin{tabular}{|c|c||c|c|c|c|c|c|c|c|c|c|c|c|c|c|}
      \hline
      &    & \multicolumn{14}{c|}{Success Rate [\%]} \\
      \cline{3-16}
      &    & \multicolumn{12}{c|}{Inclination [deg]} & & \\
      \cline{3-14}
      Model & Reference line & $-$30 & $-$20 & $-$10 & 0* & 10 & 20* & 30 & 40* & 50 & 60 & 70 & 80 & Subtotal & Total\\
      \hline
      & A& 0& 100& 100& 100& 100& 100& 100& 100& 100& 100& 100& 100& 91.7& \\
      S2S-w/o-AR & B**& 0& 100& 100& 100& 100& 100& 100& 100& 100& 100& 100& 100& 91.7& 88.0 (95/108)\\
      & C& 0& 0& 100& 100& 100& 100& 100& 100& 100& 100& 100& 66.7& 81.0& \\
      \hline
      & A& 0& 100& 100& 100& 100& 100& 100& 100& 100& 100& 100& 100& 91.7& \\
      S2S-AR & B**& 0& 100& 100& 100& 100& 100& 100& 100& 100& 100& 100& 100& 91.7& 93.5 (101/108)\\
      & C& 100& 100& 100& 100& 100& 100& 100& 100& 100& 100& 100& 66.7& 97.2& \\
      \hline
      & A& 0& 0& 0& 66.7& 100& 100& 100& 100& 100& 100& 100& 66.7& 69.4& \\
      S2M-w/o-AR & B**& 0& 0& 0& 100& 100& 100& 100& 100& 100& 100& 66.7& 0& 63.9& 66.7 (72/108)\\
      & C& 0& 0& 0& 100& 100& 100& 100& 100& 100& 100& 100& 0& 66.7& \\
      \hline
      & A& 0& 0& 100& 100& 100& 100& 100& 100& 0& 100& 100& 100& 75.0& \\
      SM2SM-w/o-AR & B**& 0& 0& 100& 100& 100& 100& 100& 100& 100& 100& 100& 100& 83.3& 81.0 (87/108)\\
      & C& 0& 0& 100& 100& 100& 100& 100& 100& 100& 100& 100& 100& 83.3& \\
      \hline
      & A& 100& 100& 100& 100& 100& 100& 100& 100& 100& 100& 100& 100& {\bf 100}& \\
      SM2SM-AR & B**& 100& 100& 100& 100& 100& 100& 100& 100& 100& 100& 100& 100& {\bf 100}& {\bf 100} (108/108)\\
      (Proposed method)& C& 100& 100& 100& 100& 100& 100& 100& 100& 100& 100& 100& 100& {\bf 100}& \\
      \hline
    \end{tabular}
    \begin{tablenotes}[para, flushleft]
      \item *: Learned inclination of the ruler
      \item **: Learned position of the ruler
    \end{tablenotes}
  \end{threeparttable}
  \end{center}
\end{table*}
The success rates of each model are shown in Table~\ref{table:result_ruler}.
First, when the models without autoregressive learning were compared, S2S-w/o-AR had a higher success rate than S2M-w/o-AR and SM2SM-w/o-AR.
As mentioned in Section~\ref{sec:method}, the S2M model was more suitable than the S2S model for IL, including for fast motion with delays.
However, this task was not particularly fast.
In addition, during the drawing task, the motion of the slave was restrained by the ruler.
The dataset of the slave's response values was easy to learn because it did not include large fluctuations, as shown in Fig.~\ref{fig:ruler_consideration}.
By contrast, fluctuations may have been contained in the master responses because the master was not restrained by anything, as shown in Fig.~\ref{fig:ruler_consideration}.
Therefore, when the master responses were used in the input or output of the model, such as in the S2M and SM2SM models, learning was difficult.
In addition, SM2SM-w/o-AR exhibited a higher performance than S2M-w/o-AR.
As described in Section~\ref{subsec:SM2SM}, the structure of the SM2SM model was more suitable than that of the S2M model because accurately understanding the relationship between the master and slave was necessary for bilateral control-based IL.
\par
Furthermore, SM2SM-AR had a higher success rate than SM2SM-w/o-AR as well as the highest success rate among all models.
In particular, compared to other methods, SM2SM-AR had a high adaptability to changes in the ruler's position and extrapolation inclinations.
As described in Section~\ref{sec:nn model}, autoregressive learning is a method that was evaluated not by the prediction error of only a single step, but by the prediction errors of all consecutive steps.
Therefore, the model could properly generate a series of motions to perform a task even for unknown environments, and the effects of the fluctuation of the master's responses were negligible.
These results indicate that the proposed model's structure and autoregressive learning improved the generalization performance for unknown environments, even with fluctuating responses.
The generalization of the proposed method, which can achieve high success rates even in unknown environments, is expected to be applied to other tasks.

\subsection{Experiment 2 (Erasing a line using an eraser)}
\label{subsec:experiment2}
\Figure[t](topskip=0pt, botskip=0pt, midskip=0pt)[width=5cm]{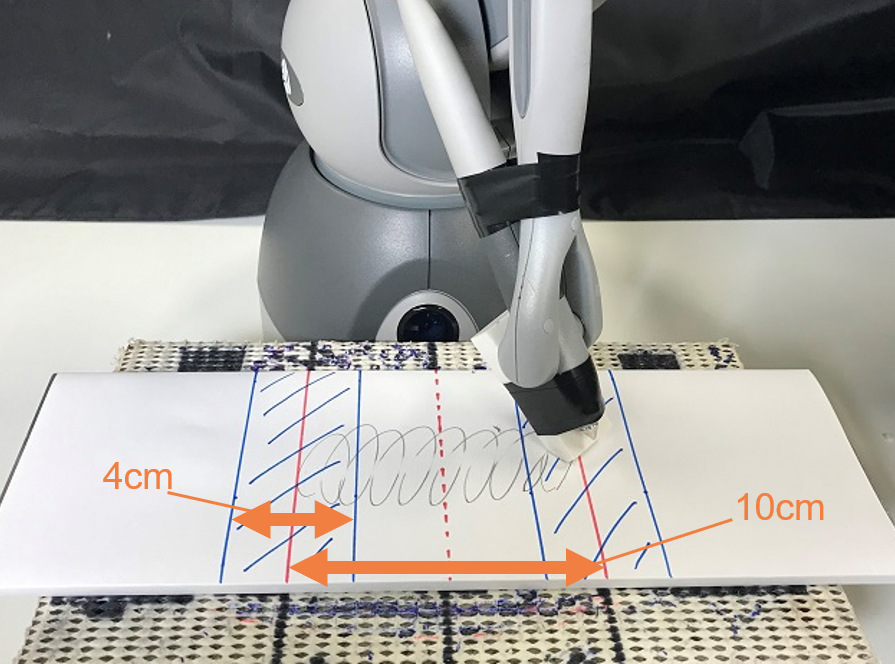}
{Experimental setup of the erasing task.
During human demonstrations of tasks, training data were collected for the task of erasing the area inside the solid red line shown in the figure.
In verifying the autonomous operation, success was defined as the case in which the movement was switched in the opposite direction in the area indicated by the blue diagonal line.
\label{fig:setup2}}
\subsubsection{Task design}
\label{subsubsec:task2}

Fig.~\ref{fig:setup2} shows the steps of this experiment.
An eraser was fixed to the slave.
In the initial condition, the eraser was in contact with the paper at the solid red line on the left shown in Fig.~\ref{fig:setup2}.
Then, the slave moved horizontally and erased a line written on the paper with the eraser.
The goal of this task was to erase a line according to various paper heights.
Adaptability to unknown paper heights was required.
To succeed in this task, proper adjustment of the contact force between the eraser and paper was required.
In this task, the robot had to operate quickly and utilize the inertial force, as considerable friction occurred between the eraser and paper.

\subsubsection{Human demonstrations and dataset for learning}
\label{subsubsec:dataset2}
We collected data with paper heights of 35, 55, and 75~mm.
Five trials were conducted for each paper height, and the total number of trials was 15.
The time for one trial was 10~s.
The dataset was collected to erase the area inside the solid red line shown in Fig.~\ref{fig:setup2}.
In the figure, the slave moved horizontally and repeatedly in the opposite direction at the solid red line.
The slave robot was teleoperated to reciprocate left and right within the area at approximately constant cycles.

\subsubsection{Neural network architecture}
\label{subsubsec:nn2}
The NN model consisted of two or four LSTM layers, followed by a fully connected layer.
During this task, two types of NN architectures were used because the robot behavior differed depending on the number of LSTM layers, and the different architectures affected the results.
The unit size of all layers was 50.
The mini-batch consisted of 100 random sets of 300 time-sequential samples corresponding to 6~s.

\subsubsection{Task Validation}
\label{subsubsec:validation2}
In verifying the autonomous operations, the performances for paper heights of 35, 45, 55, 65, and 75~mm were validated.
The paper heights of 45 and 65~mm were the untrained heights.
Success was defined as the case in which the robot erased the line within the specified area.
We defined the area of success to exclude cases in which the robot movements were too narrow or too wide as compared to successful demonstrations.
In Fig.~\ref{fig:setup2}, when the robot reciprocated to the left and right, the success was defined as the case in which the movement was switched to the opposite direction in the area indicated by the blue diagonal line.
The ability to erase the line with the appropriate force according to changes in height was an essential condition for success.
The robot executed the task for 8 s during each trial, and the case in which the robot stopped during the trial was defined as a failure.
Verification was conducted through three trials for each paper height.

\subsubsection{Experimental results}
\label{subsubsec:result2}
\begin{table*}[t]
  \begin{center}
    \begin{threeparttable}[t]
    \caption{Success rates of the erasing task}
    \label{table:result_eraser}
    \begin{tabular}{|c|c||c|c|c|c|c|c|c|}
      \hline
      & & \multicolumn{6}{c|}{Success rate based on the success area [\%]} & Total rate of robot \\
      \cline{3-8}
      &The number of& \multicolumn{5}{c|}{Height [mm]} &  & continued to perform the task \\
      \cline{3-7}
      Model & LSTM layer & 35* & 45 & 55* & 65 & 75* & Total & during the trial [\%]\\
      \hline
      & 2& 0& 0& 0& 66.7& 33.3& 20.0 (3/15) & 40.0 (6/15)\\
      S2S-w/o-AR & 4& 0& 0& 0& 100& 100& 40.0 (6/15) & {\bf 100} (15/15)\\
      \hline
      & 2& 33.3& 0& 0& 0& 0& 6.67 (1/15) & 80.0 (12/15)\\
      S2S-AR & 4& 0& 0& 0& 0& 0& 0 (0/15) & {\bf 100} (15/15)\\
      \hline
      & 2& 100& 100& 66.7& 100& 100& {\bf 93.3} (14/15) & {\bf 100} (15/15)\\
      S2M-w/o-AR & 4&  100& 66.7& 66.7& 100& 100& 86.7 (13/15) & {\bf 100} (15/15)\\
      \hline
      & 2& 0& 0& 0& 0& 33.3& 6.67 (1/15) & {\bf 100} (15/15)\\
      SM2SM-w/o-AR & 4& 100& 66.7& 100& 100& 100& {\bf 93.3} (14/15) & {\bf 100} (15/15)\\
      \hline
      SM2SM-AR & 2& 100& 100& 100& 66.7& 100& {\bf 93.3} (14/15) & {\bf 100} (15/15)\\
      (Proposed method)& 4& 100& 100& 33.3& 66.7& 100& 80.0 (12/15) & {\bf 100} (15/15)\\
      \hline
    \end{tabular}
    \begin{tablenotes}[para, flushleft]
      *: Learned height
    \end{tablenotes}
  \end{threeparttable}
  \end{center}
\end{table*}
The success rates of each model are shown in Table~\ref{table:result_eraser}.
The rates in the rightmost column of the table were determined using a different evaluation criterion than the aforementioned.
These results show the percentages of trials in which the robot could continue erasing the line without stopping, regardless of whether the success criteria were satisfied based on the success area previously described.
During this experiment, the results differed with the number of LSTM layers.
The performance was validated by changing the number of LSTM layers of each model.
As the table shows, the S2S model generally had low success rates.
Many of the failures were cases in which the robot stopped due to friction between the eraser and paper, or the robot went outside the workspace.
During this task, the robot had to move extremely quickly.
In addition, considerable friction occurred between the eraser and paper.
Thus, control delays and delays due to physical interactions occurred during the demonstrations.
The training data and results are shown in Fig.~\ref{fig:eraser_consideration}.
We focused on $\theta_{1}^{res}$ because joint 1 moved mainly during the erasing task.
In addition, the bottom of the figure shows $\tau_{2}^{res}$, which was the force exerted primarily in the vertically downward direction.
Fig.~\ref{fig:eraser_consideration}(a) shows that a delay between the response values of the master and slave occurred.
Therefore, the skill required to compensate for the delays performed by humans during the demonstrations had to be reproduced during the autonomous operation.
The results of the S2S-AR model, as an example of task failure, are shown in the center of the figure.
In the middle of the operation, $\tau_{1}^{res}$ and $\tau_{2}^{res}$ became greater than the appropriate force, and the robot could not move.
The S2S model lost this compensation skill and could not complete this task requiring fast motion.
\par
By contrast, both the S2M and SM2SM models showed high success rates.
The SM2SM-AR model could reproduce the fast motion with delays while maintaining the same level of appropriate force as the training data.
In addition, the robot could properly erase the line without stopping during all trials.
The robot applied the appropriate force even at unlearned heights.
Most failures occurred when the movements were slightly beyond the success area.
Although none of the models exhibited a perfect performance, as the success area was strictly defined, the S2M and SM2SM models achieved excellent performances in realizing fast motion while maintaining the proper force.
Because this task involved a reciprocating motion with a short cycle, a long-term prediction was not required, and it was a relatively easy task for bilateral control-based IL.
Therefore, even the conventional S2M model without autoregressive learning showed as high a success rate as that of the proposed model.
It was confirmed that the proposed method with autoregressive learning achieved an excellent performance even for a short-cycle task without adverse effects.
\Figure[t](topskip=0pt, botskip=0pt, midskip=0pt)[width=8.5cm]{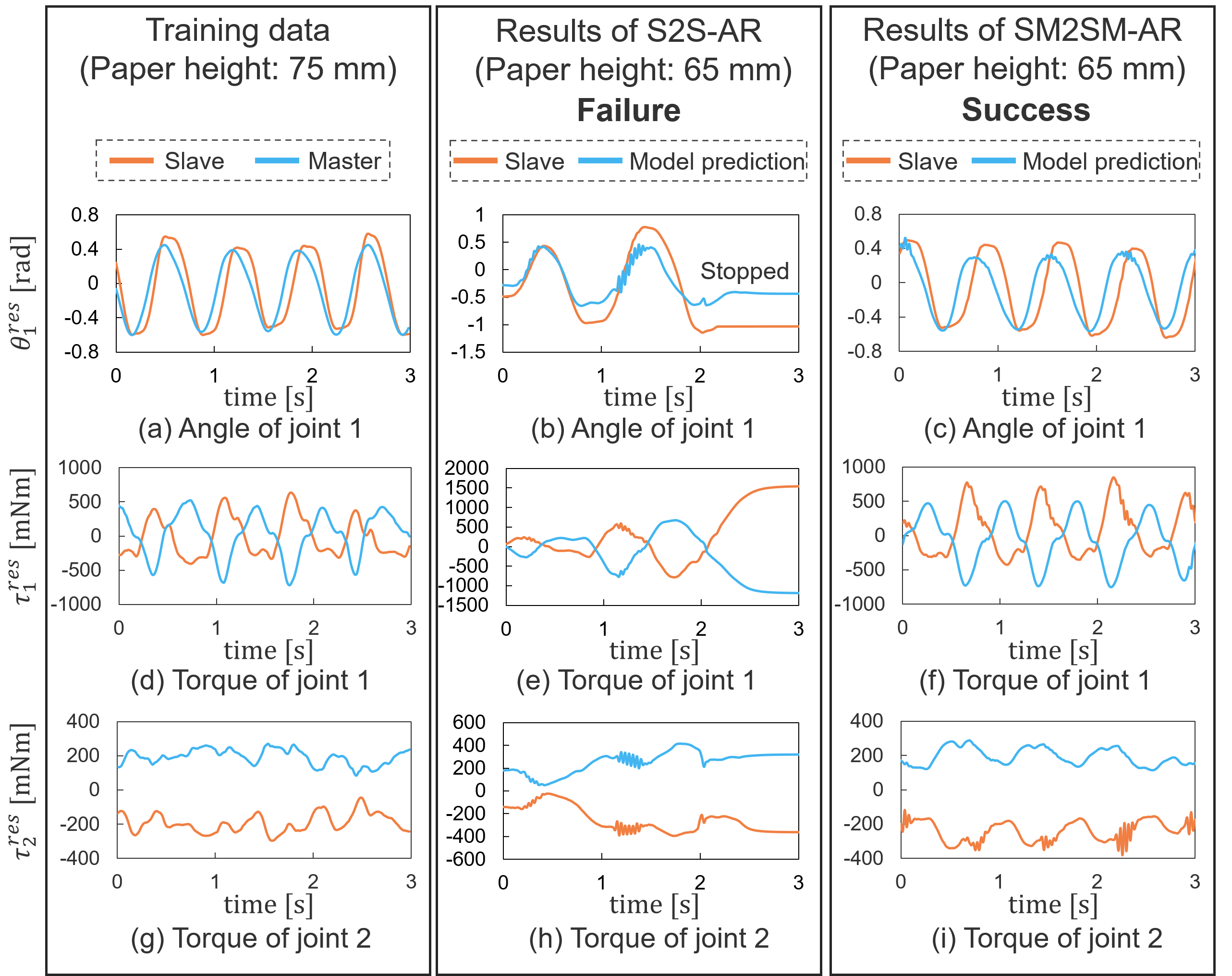}
{Training data and results of the erasing task.
The results of the S2S-AR and SM2SM-AR are shown as examples.\label{fig:eraser_consideration}}
\begin{table}[t]
  \begin{center}
    \caption{Success rates of the writing task}
    \label{table:result_ab}
    \scriptsize
    \begin{tabular}{|c||c|c|c|}
      \hline
            &\multicolumn{3}{c|}{Success Rate [\%]} \\
      \cline{2-4}
      Model & Letter ``A'' & Letter ``B'' & Total  \\
      \hline
      S2S-w/o-AR & 0 (0/4) & 0 (0/4) & 0 (0/8)\\
      \hline
      S2S-AR & 0 (0/4)& 0 (0/4)& 0 (0/8)\\
      \hline
      S2M-w/o-AR & 75.0 (3/4)& 75.0 (3/4) & 75.0 (6/8)\\
      \hline
      SM2SM-w/o-AR & 75.0 (3/4) & {\bf 100} (4/4) & 87.5 (7/8)\\
      \hline
      SM2SM-AR (proposed method)& {\bf 100} (4/4) & {\bf 100} (4/4) & {\bf 100} (8/8)\\
      \hline
    \end{tabular}
  \end{center}
\end{table}
\Figure[t](topskip=0pt, botskip=0pt, midskip=0pt)[width=18cm]{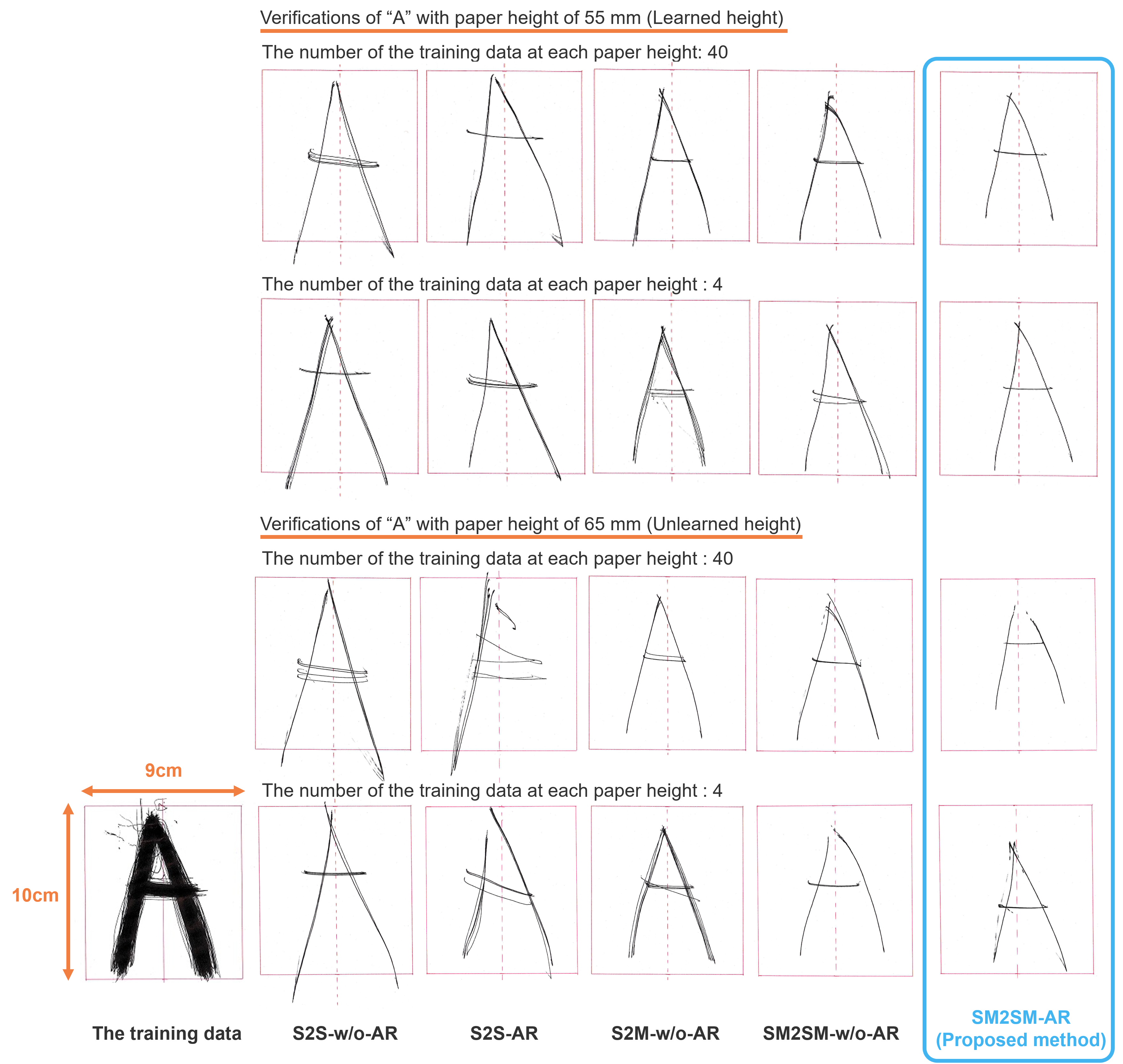}
{Training data and the results of the task of writing the letter ``A.''
The figures on the far left are the letters written by a human during the demonstrations.
Those from the training data appeared as thick because they were written on a single sheet of paper through all trials.
\label{fig:writing_considerationA}}
\Figure[t](topskip=0pt, botskip=0pt, midskip=0pt)[width=18cm]{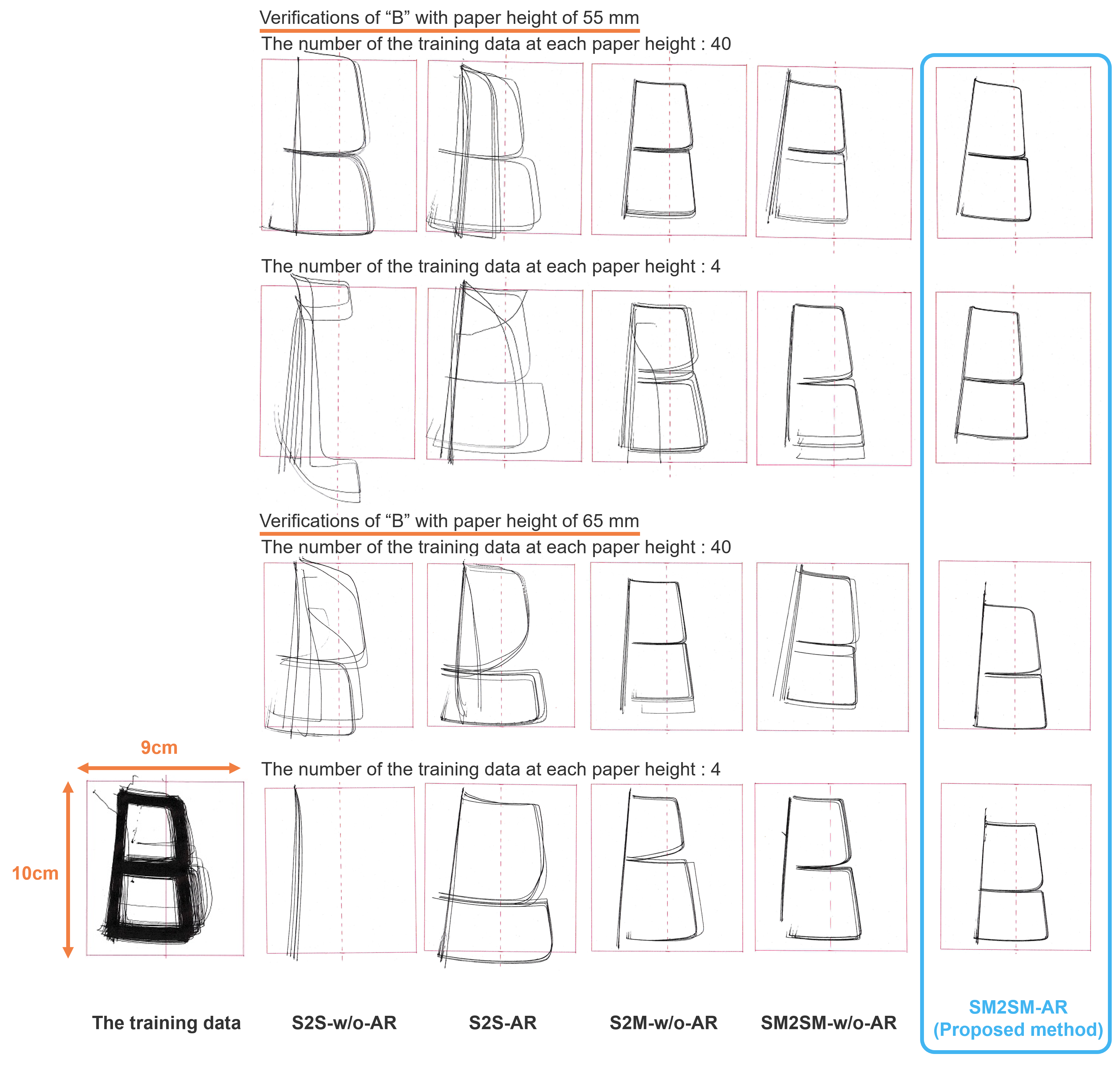}
{Training data and the results of the task of writing the letter ``B.''
\label{fig:writing_considerationB}}

\subsection{Experiment 3 (Writing letters)}
\label{subsec:experiment3}

\subsubsection{Task design}
\label{subsubsec:task3}
A ballpoint pen was fixed to the slave.
The goal of this task was to write the letters ``A'' and ``B'' on paper.
The initial position was the start point of the first stroke of each letter.
The letter was written according to specified stroke order, and then the robot went back to the first stroke again and wrote it repeatedly.
Compared to the erasing task, this writing task was a long-term operation and required a correct long-term prediction.
To succeed in this task, reproducing the proper force between the paper and pen was necessary.
In addition, the robot had to reproduce the stroke order learned from the human demonstrations. In other words, the ability to generate correct behavior based on past and current states was necessary.

\subsubsection{Human demonstrations and dataset for learning}
\label{subsubsec:dataset3}
We collected data with paper heights of 35, 55, and 75~mm.
The letters ``A'' and ``B'' were collected as separate trials.
One 15-s trial included writing the same letter four times in a row.
In other words, the number of training data in one trial was four.
Ten trials were conducted for each paper height.
In total, 120 sets of training data were collected for each letter (40 sets of training data $\times$ three heights).
The letters were written inside the solid red lines shown in Figs.~\ref{fig:writing_considerationA} and \ref{fig:writing_considerationB}.
The letters were written by humans in demonstrations such that the shape would be roughly the same during all trials without using any restraining tools including a ruler.

\subsubsection{Neural network architecture}
\label{subsubsec:nn3}
The NN model consisted of six LSTM layers, followed by a fully connected layer.
A unit size of 50 was used for all layers.
The mini-batch consisted of 100 random sets of 200 time-sequential samples corresponding to 4~s.

\subsubsection{Validation of the task}
\label{subsubsec:validation3}
For autonomous operations, the performances for paper heights of 55 and 65~mm were verified.
In addition, verification was conducted for the cases in which 4 and 40 sets of training data were used for each paper height.
Success was defined as cases in which the robot wrote the letter five times continuously inside the solid red lines shown in Figs.~\ref{fig:writing_considerationA} and \ref{fig:writing_considerationB} and with the correct stroke order.
Verification was conducted for each height and each number of training data.
Therefore, four verifications were performed (two heights $\times$ two training datasets).

\subsubsection{Experimental results}
\label{subsubsec:result3}
The success rates of the different models are listed in Table~\ref{table:result_ab}.
Only the proposed method was successful for all validations.
The results of continuously writing the letter five times are shown in Figs.~\ref{fig:writing_considerationA} and \ref{fig:writing_considerationB}.
With the conventional methods, the trajectory of the letters was unstable every time.
By contrast, with the proposed method, the letters were written with the same trajectory every time.
In particular, in the case in which the training data were few, the difference from the other methods was noticeable.
Only the proposed method could generate a trajectory with little fluctuation.
This result indicated that the proposed method generated motion with little fluctuation in the long-term thanks to autoregressive learning, {\it i.e.,} the model learned to minimize the total errors of the long-term prediction.

\section{CONCLUSION}
\label{sec:conclusion}
In this study, we proposed a method of autoregressive learning for a bilateral control-based IL.
Due to the structure and autoregressive learning of the proposed model, the performance was improved compared to the conventional methods.
During the experiments, three types of tasks were performed, and the proposed method had the highest success rate.
In addition, the proposed method improved the generalization for unknown environments not included in the training dataset.
\par
However, the proposed method must be improved in terms of model structure.
In the proposed SM2SM model, the master state predicted by the model in the previous step was input to the model during an autonomous operation, and thus the master state used in the input could be regarded as a virtual master state.
This is because the slave executes tasks alone during the autonomous operations.
If sudden environmental changes occur during the execution of the task, this state of the virtual master is likely to differ from the that of the actual master.
For example, a sudden environmental change is a situation in which the paper height is changed during the operation of the writing task.
This error affects the prediction of the model, and therefore this issue has to be solved in the future.
We plan to further improve the performance of autoregressive learning for bilateral control-based IL by changing the structure of the SM2SM model or implementing systems to correct the error.

\begin{IEEEbiography}[{\includegraphics[width=1in,height=1.25in,clip,keepaspectratio]{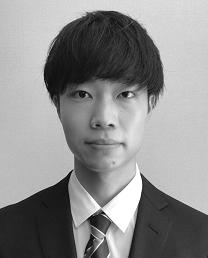}}]{Ayumu Sasagawa} (Student Member, IEEE)
  received the B.E. degree in electrical and electronic systems engineering from Saitama University, Saitama, Japan, in 2019, where he is currently pursuing the M.E. degree with the Department of Electrical and Electronic Systems. His research interests include robotics, motion generation, and neural networks.
\end{IEEEbiography}
\begin{IEEEbiography}[{\includegraphics[width=1in,height=1.25in,clip,keepaspectratio]{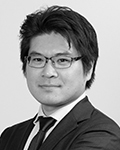}}]{Sho Sakaino} (Member, IEEE)
  received his B.E. degree in system design engineering and M.E. and Ph.D. degrees in integrated design engineering from Keio University, Yokohama, Japan, in 2006, 2008, and 2011 respectively. He was an assistant professor at Saitama University from 2011, to 2019. Since 2019, he has been an associate professor at University of Tsukuba. His research interests include mechatronics, motion control, robotics, and haptics. He received the IEEJ Industry Application Society Distinguished Transaction Paper Award in 2011 and 2020. He also received the RSJ Advanced Robotics Excellent Paper Award in 2020.
\end{IEEEbiography}
\begin{IEEEbiography}[{\includegraphics[width=1in,height=1.25in,clip,keepaspectratio]{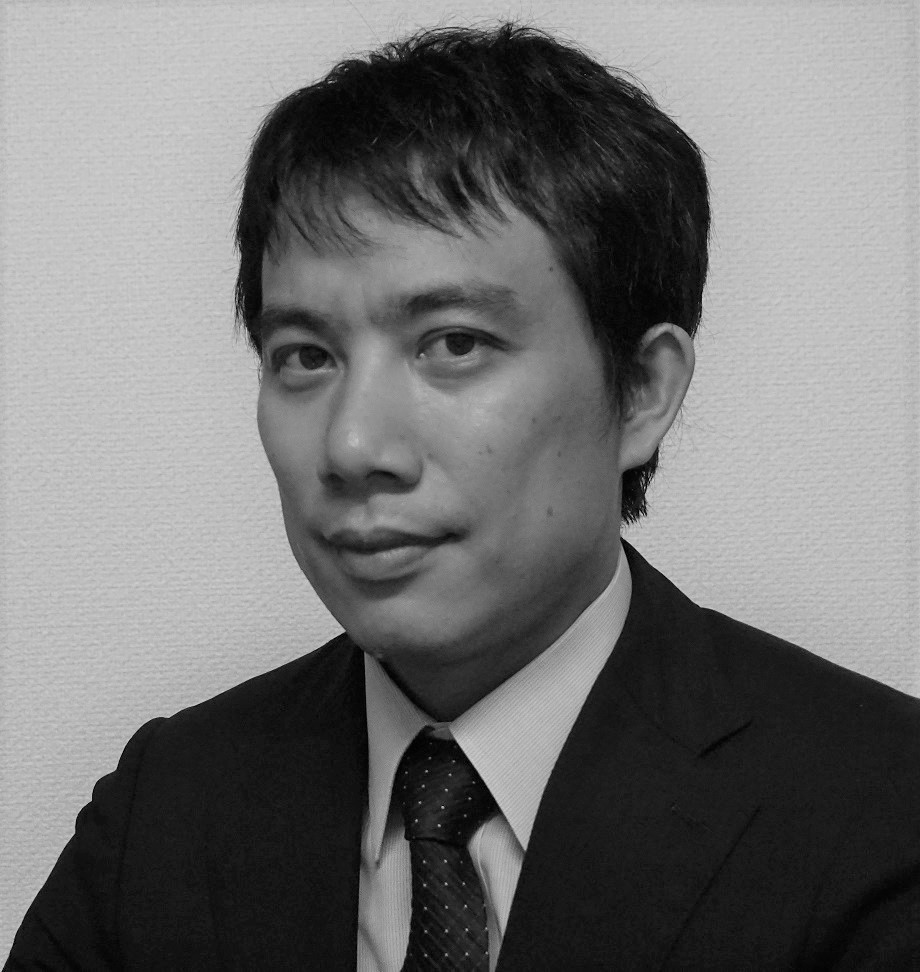}}]{Toshiaki Tsuji} (Senior Member, IEEE)
  received his B.E. degree in system design engineering and M.E. and Ph.D. degrees in integrated design engineering from Keio University, Yokohama, Japan, in 2001, 2003, and 2006, respectively. He was a Research Associate in the Department of Mechanical Engineering, Tokyo University of Science, from 2006 to 2007. He is currently an Associate Professor in the Department of Electrical and Electronic Systems, Saitama University, Saitama, Japan. His research interests include motion control, haptics and rehabilitation robots. Dr. Tsuji received the FANUC FA and Robot Foundation Original Paper Award in 2007 and 2008, respectively. He also received the RSJ Advanced Robotics Excellent Paper Award and the IEEJ Industry Application Society Distinguished Transaction Paper Award in 2020.
\end{IEEEbiography}

\EOD


\begin{thebibliography}{99}
  \bibitem{object detection} L. Jiao, F. Zhang, F. Liu, S. Yang, L. Li, Z. Feng, and R. Qu, ``A Survey of Deep Learning-Based Object Detection,'' {\it IEEE Access,} vol.~7, pp.~128837--128868, 2019.
  
  \bibitem{attention} A. Vaswani, N. Shazeer, N. Parmar, J. Uszkoreit, L. Jones, A. N. Gomez, Ł. Kaiser, and I. Polosukhin, ``Attention is all you deed,'' {\it Advances in neural information processing systems,} vol.~30, pp.~5998--6008, 2017.
  
  \bibitem{reviewer2-1} Y. Zong, T. Zhang, W. Zheng, X. Hong, C. Tang, Z. Cui, and G. Zhao, ``Cross-database micro-expression recognition: A benchmark,'' {\it IEEE Transactions on Knowledge and Data Engineering,} 2020.
  
  \bibitem{human activity recognition} W. Qi, H. Su, and A. Aliverti, ``A smartphone-based adaptive recognition and real-time monitoring system for human activities,'' {\it IEEE Transactions on Human-Machine Systems,} vol.~50, no.~5, pp.~414--423, 2020.

  \bibitem{reviewer2-2} D. Yu, C. L. P. Chen, and H. Xu, ``Intelligent decision making and bionic movement control of self-organized swarm,'' {\it IEEE Transactions on Industrial Electronics,} 2020.
  
  \bibitem{reviewer2-3} H. Huang, T. Zhang, C. Yang, and C. L. P. Chen, ``Motor learning and generalization using broad learning adaptive neural control,'' {\it IEEE Transactions on Industrial Electronics,} vol.~67, no.~10, pp.~8608--8617, 2020.

  \bibitem{remote center of motion} H. Su, Y. Hu, H. R. Karimi, A. Knoll, G. Ferrigno, and E. D. Momi, ``Improved recurrent neural network-based manipulator control with remote center of motion constraints: Experimental results,'' {\it Neural Networks,} vol.~131, pp.~291--299, 2020.

  \bibitem{human-like} H. Su, W. Qi, Y. Hu, H. R. Karimi, G. Ferrigno, and E. D. Momi, ``An incremental learning framework for human-like redundancy optimization of anthropomorphic manipulators,'' {\it IEEE Transactions on Industrial Informatics,} 2020.

  \bibitem{end-to-end} R. Rahmatizadeh, P. Abolghasemi, L. B\"{o}l\"{o}i, and S. Levine, ``Vision-based multi-task manipulation for inexpensive robots using end-to-end learning from demonstration,'' in {\it Proceedings of IEEE International Conference on Robotics and Automation (ICRA),} pp.~3758--3765, 2018.
  
  \bibitem{google} S. Levine, P. Pastor, A. Krizhevsky, and D. Quillen, ``Learning hand-eye coordination for robotic grasping with deep learning and large-scale data collection,'' {\it The International Journal of Robotics Research,} vol.~37, no.~4--5, pp.~421--436, 2018.
  \bibitem{machine learning1} T. Zhang, Z. McCarthy, O. Jow, D. Lee, X. Chen, K. Goldberg, and P. Abbeel, ``Deep imitation learning for complex manipulation tasks from virtual reality teleoperation,'' in {\it Proceedings of IEEE International Conference on Robotics and Automation (ICRA),} pp.~1--8, 2018.
  
  \bibitem{poke} P. Agrawal, A. Nair, P. Abbeel, J. Malik, and S. Levine, ``Learning to poke by poking: experiential learning of intuitive physics,'' in {\it Proceedings of Advances in neural information processing systems,} pp.~5074--5082, 2016.
  \bibitem{imitation learning survey2} H. Ravichandar, A. S. Polydoros, S. Chernova, and A. Billard, ``Recent advances in robot learning from demonstration,'' {\it Annual Review of Control, Robotics, and Autonomous Systems,} vol.~3, pp.~297--330, 2020.
  \bibitem{imitation learning survey} B. Fang, S. Jia, D. Guo, M. Xu, S. Wen, and F. Sun, ``Survey of imitation learning for robotic manipulation,'' {\it International Journal of Intelligent Robotics and Applications,} no.~3, pp.~362--369, 2019.
  
  \bibitem{one-shot} T. Yu, C. Finn, A. Xie, S. Dasari, T. Zhang, P. Abbeel, and S. Levine, ``One-shot imitation from observing humans via domain-adaptive meta-learning,'' in {\it Proceedings of Robotics: Science and Systems (RSS),} 2018.
  
  \bibitem{cooperation human-human demo} D. Vogt, S. Stepputtis, S. Grehl, B. Jung, and H. B. Amor, ``A system for learning continuous human-robot interactions from human-human demonstrations,'' in {\it Proceedings of IEEE International Conference on Robotics and Automation (ICRA),} pp.~2882--2889, 2017.
  
  \bibitem{ogata} P. C. Yang, K. Sasaki, K. Suzuki, K. Kase, S. Sugano, and T. Ogata, ``Repeatable folding task by humanoid robot worker using deep learning,'' {\it IEEE Robotics and Automation Letters,} vol.~2, no.~2, pp.~397--403, 2017.
  
  \bibitem{relay policy} A. Gupta, V. Kumar, C. Lynch, S. Levine, and K. Hausman, ``Relay policy learning: Solving long-horizon tasks via imitation and reinforcement learning,'' arXiv:1910.11956, 2019.

  \bibitem{iit1} P. Kormushev, S. Calinon, and D. G. Caldwell, ``Imitation learning of positional and force skills demonstrated via kinesthetic teaching and haptic input,'' {\it Advanced Robotics,} vol.~25, no.~5, pp.~581--603, 2011.
  
  \bibitem{iit2} L. Rozo, P. Jim\'enez, and C. Torras, ``A robot learning from demonstration framework to perform force-based manipulation tasks,'' {\it Intel Serv Robotics,} vol.~6, no.~1, pp.~33--51, 2013.
  
  \bibitem{cooperationIIT} L. Rozo, D. Bruno, S. Calinon, and D. G. Caldwell, ``Learning optimal controllers in human-robot cooperative transportation tasks with position and force constraints,'' in {\it Proceedings of IEEE/RSJ International Conference on Intelligent Robots and Systems (IROS),} pp.~1024--1030, 2015.

  \bibitem{force} A. X. Lee, H. Lu, A. Gupta, S. Levine, and P. Abbeel, ``Learning force-based manipulation of deformable objects from multiple demonstrations,'' in {\it Proceedings of IEEE International Conference on Robotics and Automation (ICRA),} pp.~177--184, 2015.
  
  \bibitem{harada} H. Ochi, W. Wan, Y. Yang, N. Yamanobe, J. Pan, and K. Harada, ``Deep learning scooping motion using bilateral teleoperations,'' in {\it Proceedings of 2018 3rd International Conference on Advanced Robotics and Mechatronics (ICARM),} pp.~118--123, 2018.

  \bibitem{force whiteboard} P. Kormushev, D. N. Nenchev, S. Calinon, and D. G. Caldwell, ``Upper-body kinesthetic teaching of a free-standing humanoid robot,'' in {\it Proceedings of IEEE International Conference on Robotics and Automation (ICRA),} pp.~3970--3975, 2011.

  \bibitem{force writing} M. Tykal, A. Montebelli, and V. Kyrki, ``Incrementally assisted kinesthetic teaching for programming by demonstration,'' in {\it Proceedings of 2016 11th ACM/IEEE International Conference on Human-Robot Interaction (HRI),} pp.~205--212, 2016.
  
  \bibitem{adachi} T. Adachi, K. Fujimoto, S. Sakaino, and T. Tsuji, ``Imitation learning for object manipulation based on position/force information using bilateral control,'' in {\it Proceedings of IEEE/RSJ International Conference on Intelligent Robots and Systems (IROS),} pp.~3648--3653, 2018.

  \bibitem{sasagawa} A. Sasagawa, K. Fujimoto, S. Sakaino, and T. Tsuji, ``Imitation learning based on bilateral control for human–robot cooperation,'' {\it IEEE Robotics and Automation Letters,} vol.~5, no.~4, pp.~6169--6176, 2020.
  
  \bibitem{bilate} S. Sakaino, T. Sato, and K. Ohnishi, ``Multi-DOF micro-macro bilateral controller using oblique coordinate control,'' {\it IEEE Transactions on Industrial Informatics,} vol.~7, no.~3, pp.~446--454, 2011.
  
  \bibitem{symmetric_bilate} T. Kitamura, N. Mizukami, H. Mizoguchi, S. Sakaino, and T. Tsuji, ``Bilateral control in the vertical direction using functional electrical stimulation,'' {\it IEEJ Journal of Industry Applications,} vol.~5, no.~5, pp.~398–-404, 2016.

  \bibitem{bilate reviewer1} Z. Chen, F. Huang, C. Yang, and B. Yao, ``Adaptive fuzzy backstepping control for stable nonlinear bilateral teleoperation manipulators with enhanced transparency performance,'' {\it IEEE Transactions on Industrial Electronics,} vol.~67, no.~1, pp.~746--756, 2019.

  \bibitem{bilate reviewer2} Z. Chen, F. Huang, W. Sun, J. Gu, and B. Yao, ``RBF-neural-network-based adaptive robust control for nonlinear bilateral teleoperation manipulators with uncertainty and time delay,'' {\it IEEE/ASME Transactions on Mechatronics,} vol.~25, no.~2, pp.~906--918, 2019.

  \bibitem{LSTM} S. Hochreiter and J. Schmidhuber, ``Long short-term memory,'' {\it Neural Computation,} vol.~9, no.~8, pp.~1735--1780, 1997.
  
  \bibitem{techer forcing} R. J. Williams and D. Zipser, ``A learning algorithm for continually running fully recurrent neural networks,'' {\it Neural computation,} vol.~1, no.~2, pp.~270--280, 1989.
  \bibitem{professor forcing} A. Lamb, A. Goyal, Y. Zhang, S. Zhang, A. Courville, and Y. Bengio ``Professor forcing: A new algorithm for training recurrent networks,'' {\it Advances in neural information processing systems,} pp.~4601--4609, 2016.
  
  \bibitem{ogata closed} K. Kase, R. Nakajo, H. Mori, and T. Ogata, ``Learning multiple sensorimotor units to complete compound tasks using an RNN with multiple attractors,'' in {\it Proceedings of IEEE/RSJ International Conference on Intelligent Robots and Systems (IROS),} 2019.
  
  \bibitem{4ch bilate 1} W. Iida and K. Ohnishi, ``Reproducibility and operationality in bilateral teleoperation,'' in {\it Proceedings of the IEEE International Workshop on Advanced Motion Control,} pp.~217--222, 2004.
  \bibitem{4ch bilate 2} K. Tanida, T. Okano, T. Murakami, and K. Ohnishi, ``Control structure determination of bilateral system based on reproducibility and operationality,'' {\it IEEJ Journal of Industry Applications,} vol.~8, no.~5, pp.~767–-778, 2019.
  
  \bibitem{dob} K. Ohnishi, M. Shibata, and T. Murakami, ``Motion control for advanced mechatronics,'' {\it IEEE/ASME Transaction on Mechatronics,} vol.~1, no.~1, pp.~56--67, 1996.
  
  \bibitem{rfob} T. Murakami, F. Yu, and K. Ohnishi, ``Torque sensorless control in multidegree-of-freedom manipulator,'' {\it IEEE Transactions on Industrial Electronics,} vol.~40, no.~2, pp.~259--265, 1993.
  
  \bibitem{yamazaki}T. Yamazaki, S. Sakaino, and T. Tsuji, ``Estimation and kinetic modeling of human arm using wearable robot arm,'' {\it Electrical Engineering in Japan,} vol.~199, no.~3, pp.~57-–67, 2017.
  
  \bibitem{adam} D. P. Kingma and J. Ba, ``Adam: A Method for Stochastic Optimization,'' in {\it Proceedings of 3rd International Conference on Learning Representations,} San Diego, CA, USA, 2015.

  \bibitem{exposure bias1} S. Bengio, O. Vinyals, N. Jaitly, and N. Shazeer, ``Scheduled sampling for sequence prediction with recurrent neural networks,'' {\it Advances in Neural Information Processing Systems,} Vol.~1, pp.~1171--1179, 2015.
  \bibitem{exposure bias2} W. Zhang, Y. Feng, F. Meng, D. You, and Q. Liu, ``Bridging the gap between training and inference for neural machine translation,'' in {\it Proceedings of the 57th Annual Meeting of the Association for Computational Linguistics,} pp.~4334–4343, 2019.
  \bibitem{exposure bias3} Y. Xu, K. Zhang, H. Dong, Y. Sun, W. Zhao, and Z. Tu, ``Rethinking exposure bias in language modeling,'' {\it arXiv:1910.11235,} 2019.
  \bibitem{exposure bias4} M. A. Ranzato, S. Chopra, M. Auli, and W. Zaremba, ``Sequence level training with recurrent neural networks,'' {\it arXiv:1511.06732}, 2015.

\end{thebibliography}
\end{document}